# Human–LLM Dialogue Improves Diagnostic Accuracy in Emergency Care

Burcu Sayin[1], Ngoc Vo Hong[2], Ipek Baris Schlicht[4], Jacopo Staiano[1], Pasquale Minervini[5], Sara Allievi[2], Nicola Susca[2], Nicola Osti[2], Alberto Maino[2,3], Vito Racanelli[2,3], Andrea Passerini[1*]

[1]Department of Information Engineering and Computer Science, University of Trento, Italy
[2]Department of Medicine, Azienda Sanitaria Universitaria Integrata del Trentino (ASUIT), Trento, Italy
[3]Center for Medical Sciences (CISMed), University of Trento, Trento, Italy
[4]Universitat Politècnica de València, Spain
[5]The University of Edinburgh, UK
*Corresponding author, e-mail: andrea.passerini@unitn.it

## Abstract

Clinical decision-making in emergency medicine demands rapid, accurate diagnoses under uncertainty. Despite benchmark progress, evidence for LLMs as interactive aids in live physician workflows remains sparse. MedSyn lets physicians iteratively query an LLM provided with the full clinical record while initially viewing only the chief complaint. Seven physicians (three seniors, four residents) completed baseline and AI-assisted sessions across 52 MIMIC-IV cases stratified by difficulty. Blinded evaluation showed residents' Hard-case correctness rose from 0.589 to 0.734; difficulty-standardised completely-correct rates confirmed a medium effect ($\Delta = 0.092$; $p = 0.071$; $d = 0.47$). Automated metrics corroborated these gains: standardised any-match accuracy improved by 0.156 ($p < 0.0001$), and residents showed the largest F1 gain ($\Delta = 0.138$; $p < 0.0001$). Dialogue analysis revealed expertise-dependent strategies (seniors asked targeted, hypothesis-driven questions; residents relied on broader queries) and cross-expertise concordance increased ($\Delta = 0.145$; $p < 0.0001$). Interactive LLM support meaningfully enhances diagnostic reasoning.

## Introduction

Clinical decision-making in emergency care is a complex, high-stakes process that requires physicians to rapidly synthesize evolving, incomplete, and often ambiguous patient information under severe time pressure[1]. In this setting, clinicians must identify potentially life-threatening conditions while managing diagnostic uncertainty, cognitive overload, emotional stress, resource constraints, and multiple patients simultaneously[2]. These demands contribute to diagnostic error, a major source of patient harm for which emergency departments account for a disproportionate share of missed or delayed diagnoses[3]. The challenge is further amplified by the growing volume and complexity of electronic health record (EHR) data, which increases information overload and detracts from efficient diagnostic reasoning[4]. Such pressures may be particularly burdensome for less experienced clinicians, who often have less developed pattern recognition and clinical intuition[5]. Although diagnostic decision support systems have been proposed to address these challenges, their adoption remains limited by poor EHR integration, lack of interoperability, and clinician distrust[6].

Large language models (LLMs) have emerged as a promising new approach for supporting clinical reasoning[7–10]. A recent systematic review and meta-analysis of 83 studies found no significant difference in diagnostic accuracy between generative AI and non-expert physicians, whereas AI performed significantly worse than expert physicians[7]. Encouraging results have also been reported in specific clinical applications: retrieval-augmented LLM workflows improved emergency department triage, referral, and diagnostic performance on 2,000 MIMIC-IV cases[8]; a two-stage prompting framework that explicitly mimics verification in physician reasoning improved diagnostic accuracy by up to 5.2% and reduced uncertainty[9]; and an uncertainty-aware system improved expert recognition of diagnostic uncertainty when used collaboratively with clinicians[10]. However, important limitations constrain direct clinical deployment. In simulated consultations, AI chatbots have shown high rates of unnecessary test ordering and medication prescribing, as well as concerning sociodemographic disparities in care quality[11]. Performance in specialized domains such as stroke care may also fall below clinical competency thresholds, and hallucinated but plausible clinical outputs remain a major patient-safety concern[12]. Together, these findings suggest that the most appropriate role for LLMs in emergency medicine is not autonomous decision-making, but structured support for physician reasoning. A recent randomised controlled trial in cardiology[13] provides complementary evidence: when nine cardiologists used an LLM assistant (AMIE) to manage 107 complex cases, diagnostic accuracy and management quality matched or exceeded unassisted specialist performance. However, that study evaluated a single-specialty, consultation-style interaction and did not examine how expertise level modulates the benefit of AI assistance.

This perspective aligns with a growing body of work on physician-AI collaboration and multi-agent reasoning. Multi-agent conversational systems have shown that structured deliberation can outperform single-pass inference, including rare-disease diagnosis frameworks that surpassed standalone GPT-4 and human-in-the-loop systems that produced more comprehensive and traceable recommendations than clinician-written outputs[14,15]. At the same time, physician benefit appears to depend not only on model capability, but also on how clinicians interact with the system, how outputs are explained, and how AI is positioned within the clinical workflow. Prior work suggests that no single input strategy reliably improves reasoning outcomes[16], whereas explanation design can substantially influence clinician uptake, confidence, and reliance on AI advice, albeit with marked inter-individual variability[17,18]. Adoption is further shaped by workflow integration, evolving perceptions of usefulness over time, and social stigma associated with using generative AI as a primary decision tool rather than as verification or support[19,20]. More broadly, although multi-LLM and multi-agent systems have been proposed to improve robustness and reasoning, evaluation has largely remained benchmark-centric and focused on closed-ended tasks[21–28]. Recent multi-turn benchmarks and collaborative evaluations show that interactive reasoning remains challenging because of retrieval failures, error propagation, limited adaptability, and instability, particularly in open-source systems[29–35]. Within healthcare, related work has examined clinical summarization, simulated patient–doctor dialogue, pathway-based evaluation, and multi-agent diagnostic systems, but much of this evidence remains simulation-driven or model-only, with persistent concerns regarding hallucinations, omissions, specialty coverage, and the limited clinical validity of standard automatic metrics such as ROUGE and BLEU[36–47]. Consequently, a key gap remains in understanding whether LLMs can meaningfully improve clinical reasoning when used by physicians in realistic, iterative diagnostic workflows. Most prior studies have focused on standalone model performance or single-turn interactions, rather than on multi-turn physician-LLM collaboration that more closely reflects clinical decision-making in practice. Evidence is also limited regarding how such support affects physicians with different levels of expertise, despite the importance of this question for both training and workforce augmentation.

This work focuses on three research questions: (1) Does the integration of LLMs into clinical workflows lead to improved diagnostic accuracy in emergency care? (2) How does LLM assistance influence the diagnostic accuracy and decision-making speed of physicians with varying levels of expertise? (3) To what extent can LLMs reduce the performance gap in diagnostic decision-making between expert and non-expert physicians? To address these questions, we developed MedSyn[48] (Fig. 1) as an experimental framework for studying interactive physician-LLM collaboration. In interactive sessions with MedSyn, the physician initially sees only the patient's chief complaint, whereas the LLM assistant has access to the full clinical record excluding the ground-truth diagnosis. The physician can iteratively query the assistant for clarification, preliminary assessment, and diagnostic suggestions. This design approximates a clinically relevant pattern of decision support in which the model informs, but does not replace, physician reasoning. We conducted a controlled evaluation in which seven physicians (three senior physicians and four residents) completed baseline (unassisted) and interactive (AI-assisted) diagnostic sessions across 52 emergency medicine cases from MIMIC-IV[49], stratified by difficulty (easy, medium, hard; see Methods: Sample Selection for stratification criteria). We assessed diagnostic accuracy using automated and clinician-judged measures, examined inter-user agreement, and analysed dialogue patterns to characterise how physicians at different expertise levels engaged with the LLM.

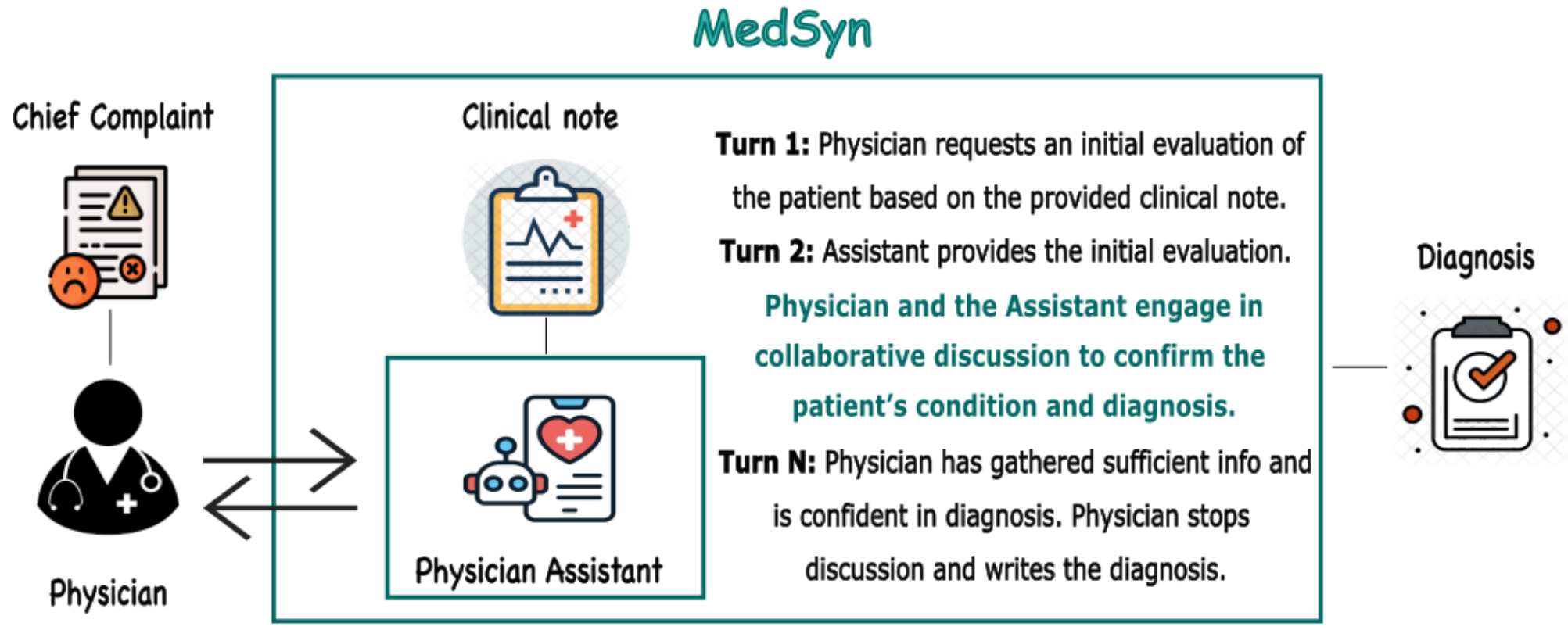

**Fig. 1 |** MedSyn Framework.

# Results

We assessed diagnostic performance using two complementary approaches. Automated fuzzy-matching metrics (any-match accuracy, exact-match accuracy, diagnosis-set F1) provide reproducible, scalable measurement of string-level concordance with discharge diagnoses and support fine-grained stratification by difficulty, session, and expertise. Blinded expert evaluation, in which a senior clinician rated each of the 364 participant-case diagnoses on a three-point ordinal scale (wrong, partially correct, completely correct), provides the clinically definitive assessment of diagnostic quality. We report automated metrics first because they support the granular stratified analyses that require larger effective sample sizes, and then present the expert assessment as clinical validation. The agreement between the two approaches is analysed explicitly in a dedicated section. To enable direct comparison, we also applied the same difficulty-standardisation and bootstrap procedure to the manual evaluation, using the most stringent criterion, the completely-correct rate, as a binary endpoint.

## Diagnostic Accuracy

We evaluated diagnostic accuracy across four sessions (two baseline and two interactive) involving seven physicians (three internal medicine experts and four residents), with each physician diagnosing 13 cases per session. To account for differences in case difficulty across sessions, all reported metrics were difficulty-standardized using fixed weights for easy (3/13), medium (6/13), and hard (4/13) cases. The primary automated outcome measure, difficulty-standardized any-match accuracy, increased significantly from baseline to interactive conditions at the cohort level (median, 0.369 vs. 0.577; mean improvement, 0.156; 95% bootstrap CI, 0.083–0.232; $p < 0.0001$; Table 1, Fig. 2). This improvement was observed in both groups: experts (mean improvement, 0.159; $p < 0.0001$) and residents (mean improvement, 0.153; $p < 0.0001$). Difficulty-standardized diagnosis-set F1 also improved significantly at the cohort level (median, 0.310 vs. 0.449; mean improvement, 0.111; 95% CI, 0.033–0.190; $p = 0.0030$). This effect was driven primarily by residents (mean improvement, 0.138; $p < 0.0001$), whereas experts showed a smaller, non-significant gain (mean improvement, 0.075; $p = 0.2945$). Difficulty-standardized exact-match accuracy improved among residents (mean improvement, 0.100; 95% CI, 0.010–0.185; $p = 0.0074$) but not among experts (mean change, -0.031; $p = 0.5159$), suggesting that LLM assistance may provide greater benefit to less experienced clinicians. Difficulty-standardized time per case did not differ significantly between conditions at the cohort level (mean change, -0.321 min; $p = 0.3057$). Residents showed a trend toward faster completion in the interactive sessions (mean change, -0.709 min; $p = 0.1015$), whereas experts showed a slight, non-significant increase (mean change, +0.196 min; $p = 0.5156$).

**Table 1 | Standardized performance: baseline vs interactive (bootstrap, 20 000 replicates)**

| Metric | Group | N | Baseline Median | Interactive Median | Mean Δ | 95% CI | p-value | d |
|---|---|---|---|---|---|---|---|---|
| **Any-match** | All | 7 | 0.369 [0.280, 0.460] | 0.577 [0.500, 0.596] | +0.156 | [0.083, 0.232] | <0.0001* | 1.25 |
| **Any-match** | Senior | 3 | 0.487 [0.428, 0.500] | 0.615 [0.596, 0.635] | +0.159 | [0.090, 0.285] | <0.0001* | 0.83 |
| **Any-match** | Resident | 4 | 0.280 [0.243, 0.346] | 0.500 [0.442, 0.519] | +0.153 | [0.046, 0.260] | <0.0001* | 0.90 |
| | | | | | | | | |
| **F1** | All | 7 | 0.310 [0.253, 0.389] | 0.449 [0.433, 0.487] | +0.111 | [0.033, 0.190] | 0.0030* | 0.85 |
| **F1** | Senior | 3 | 0.442 [0.376, 0.456] | 0.481 [0.453, 0.511] | +0.075 | [-0.017, 0.231] | 0.2945 | 0.32 |
| **F1** | Resident | 4 | 0.253 [0.212, 0.303] | 0.445 [0.385, 0.460] | +0.138 | [0.053, 0.223] | <0.0001* | 0.94 |
| | | | | | | | | |
| **Exact-match** | All | 7 | 0.212 [0.189, 0.285] | 0.308 [0.269, 0.346] | +0.044 | [-0.045, 0.130] | 0.3480 | 0.30 |

| | | | | | | | | | |
|---|---|---|---|---|---|---|---|---|---|
| **Exact-match** | Senior | 3 | 0.330 [0.271, 0.383] | 0.308 [0.269, 0.327] | -0.031 | [-0.128, 0.135] | 0.5159 | -0.12 |
| **Exact-match** | Resident | 4 | 0.189 [0.166, 0.212] | 0.327 [0.269, 0.346] | +0.100 | [0.010, 0.185] | 0.0074* | 0.76 |
| | | | | | | | | |
| **Time (min)** | All | 7 | 4.97 [3.79, 5.51] | 4.46 [3.64, 4.83] | -0.321 | [-0.935, 0.260] | 0.3057 | -0.31 |
| **Time (min)** | Senior | 3 | 2.63 [2.24, 3.79] | 3.24 [2.73, 3.90] | +0.196 | [-0.401, 0.610] | 0.5156 | 0.21 |
| **Time (min)** | Resident | 4 | 5.51 [5.16, 5.85] | 4.78 [4.35, 5.22] | -0.709 | [-1.388, 0.173] | 0.1015 | -0.53 |

* indicates statistical significance at α = 0.05. Bootstrap: 20,000 replicates, seed 42, percentile CI, two-sided p-value.

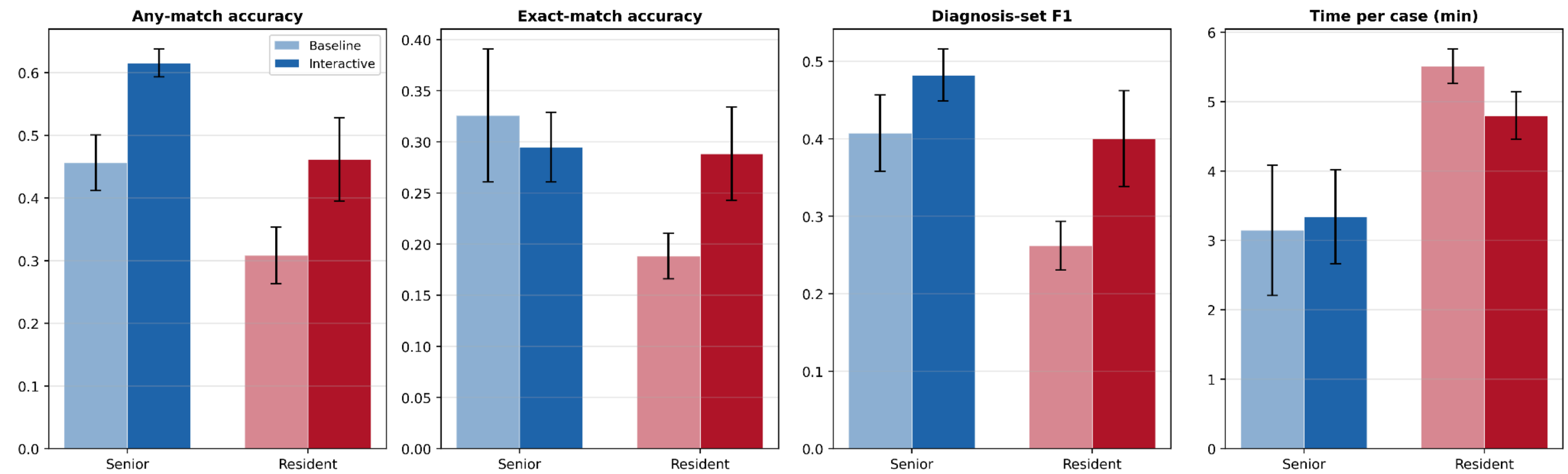

**Fig. 2 |** Four-panel grouped bar chart showing standardized any-match accuracy, exact-match accuracy, diagnosis-set F1, and time per case. Senior (blue) vs Resident (red); lighter bars = baseline, darker bars = interactive. Error bars = Standard Error of the Mean (SEM).

Difficulty-stratified analysis (Figure 3) reveals that the interactive benefit is concentrated on Medium and Hard cases. In Easy cases, performance either remained flat or declined; a striking pattern most visible in exact-match accuracy, which dropped from 0.619 to 0.167 for seniors and from 0.500 to 0.125 for residents. This counter-intuitive result likely reflects a diagnostic-expansion effect: when the LLM is available, physicians tend to broaden their differential even for straightforward presentations, adding plausible but ultimately incorrect diagnoses that hurt exact-match while still including the correct one (any-match remains stable). In hard cases, by contrast, residents' any-match accuracy more than doubled (0.214 → 0.531) and their F1 tripled (0.167 → 0.500), demonstrating that LLM assistance is most impactful when diagnostic complexity exceeds the physician's baseline capacity.

Figure 4 traces the session-by-session trajectory (S1 → S2 → S3 → S4) and reveals several patterns not visible in aggregated comparisons. First, the interactive boost is reproducible: both S2 and S4 show a clear upward step relative to the preceding baseline, with comparable magnitude. Second, S3 performance reverts to near-S1 levels for residents (any-match drops from ~0.50 in S2 back to ~0.32 in S3), arguing against a learning-effect explanation; physicians did not retain knowledge from S2 into S3, which supports the interpretation that accuracy gains require real-time LLM access. Third, the group trajectories differ in shape: the senior curve is flatter (oscillating between ~0.45 and ~0.65 on any-match) while the resident curve shows wider swings (~0.28 to ~0.50), reflecting residents' greater responsiveness to LLM availability. Finally, the exact-match panel reveals that seniors' exact-match drops sharply in S4 despite stable any-match, consistent with the diagnostic-expansion effect described above, where repeated interactive sessions encourage broader differentials at the cost of exact-match precision.

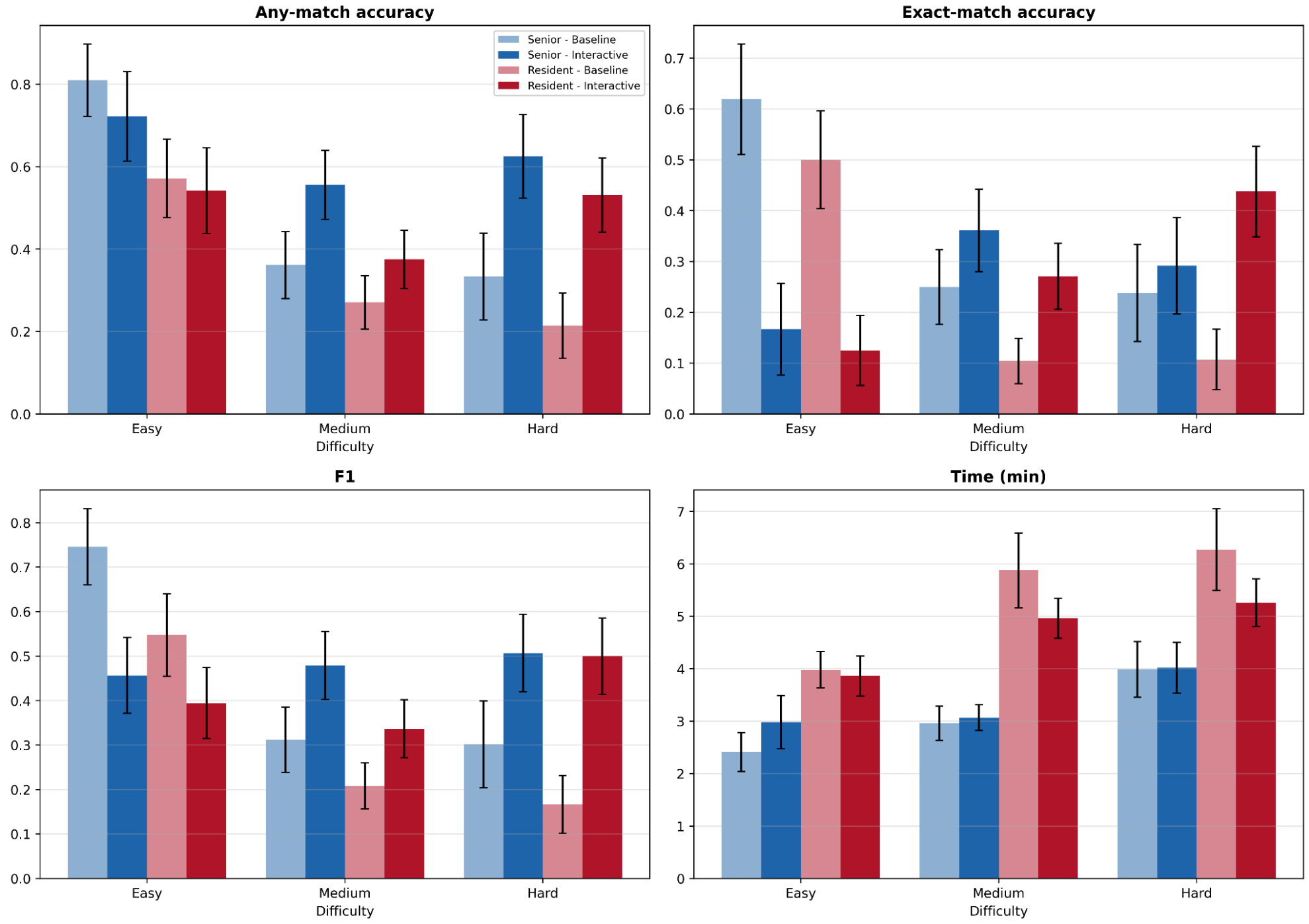

**Fig. 3 |** Four-panel grouped bar chart (any-match, exact-match, F1, time) stratified by Easy/Medium/Hard. Error bars = SEM. The interactive benefit concentrates on Medium and Hard cases; Easy-case exact-match paradoxically declines.

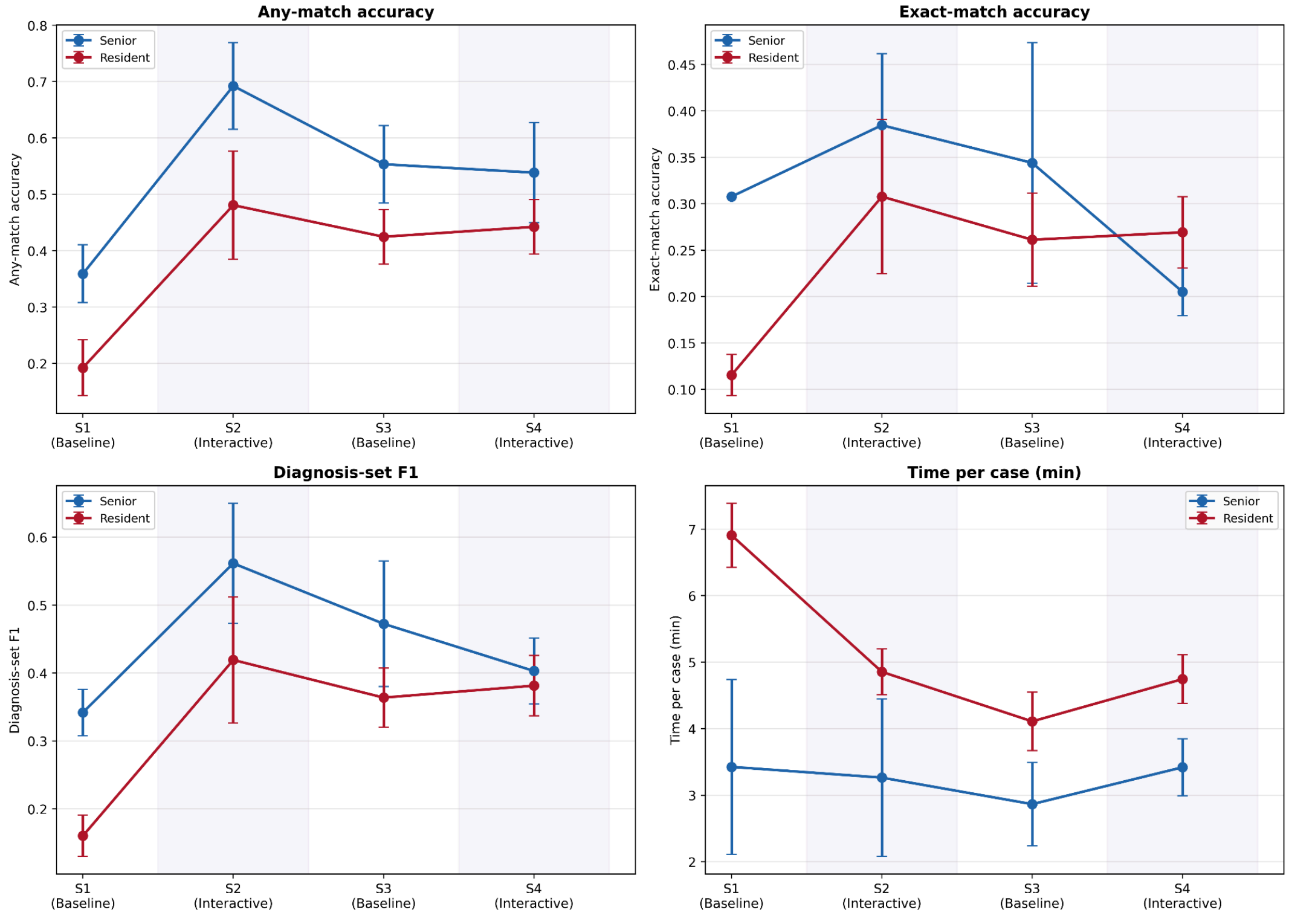

**Fig. 4 |** Four-panel line plot with error bars showing session-by-session trends for any-match, exact-match, F1, and time. Purple shading = interactive sessions. Senior (blue), Resident (red).

Figure 5 directly addresses the following research question: "To what extent can LLMs reduce the performance gap in diagnostic decision-making between expert and non-expert physicians?" by plotting the Senior-minus-Resident gap as a function of difficulty. Under baseline conditions, the expertise gap is wide for Easy cases (any-match: 0.24) and narrows toward Hard cases (0.12), reflecting the well-documented finding that difficult cases are challenging regardless of experience. With interactive support, the gap compresses across difficulty levels. On Hard cases, where equitable access to diagnostic reasoning support matters most, any-match gap shrinks from 0.12 to 0.09, and the F1 gap inverts entirely (residents slightly outperform seniors: 0.500 vs 0.507). This convergence suggests that LLM-mediated interaction compensates for experience-dependent knowledge gaps, effectively democratising access to clinical information. The blinded expert assessment corroborates this pattern: the Senior-Resident gap in mean ordinal correctness narrows from 0.316 (baseline) to 0.162 (interactive) on Hard cases and from 0.198 to 0.104 on Medium cases, while on Easy cases both groups perform near ceiling and the gap remains negligible (0.107 vs 0.118).

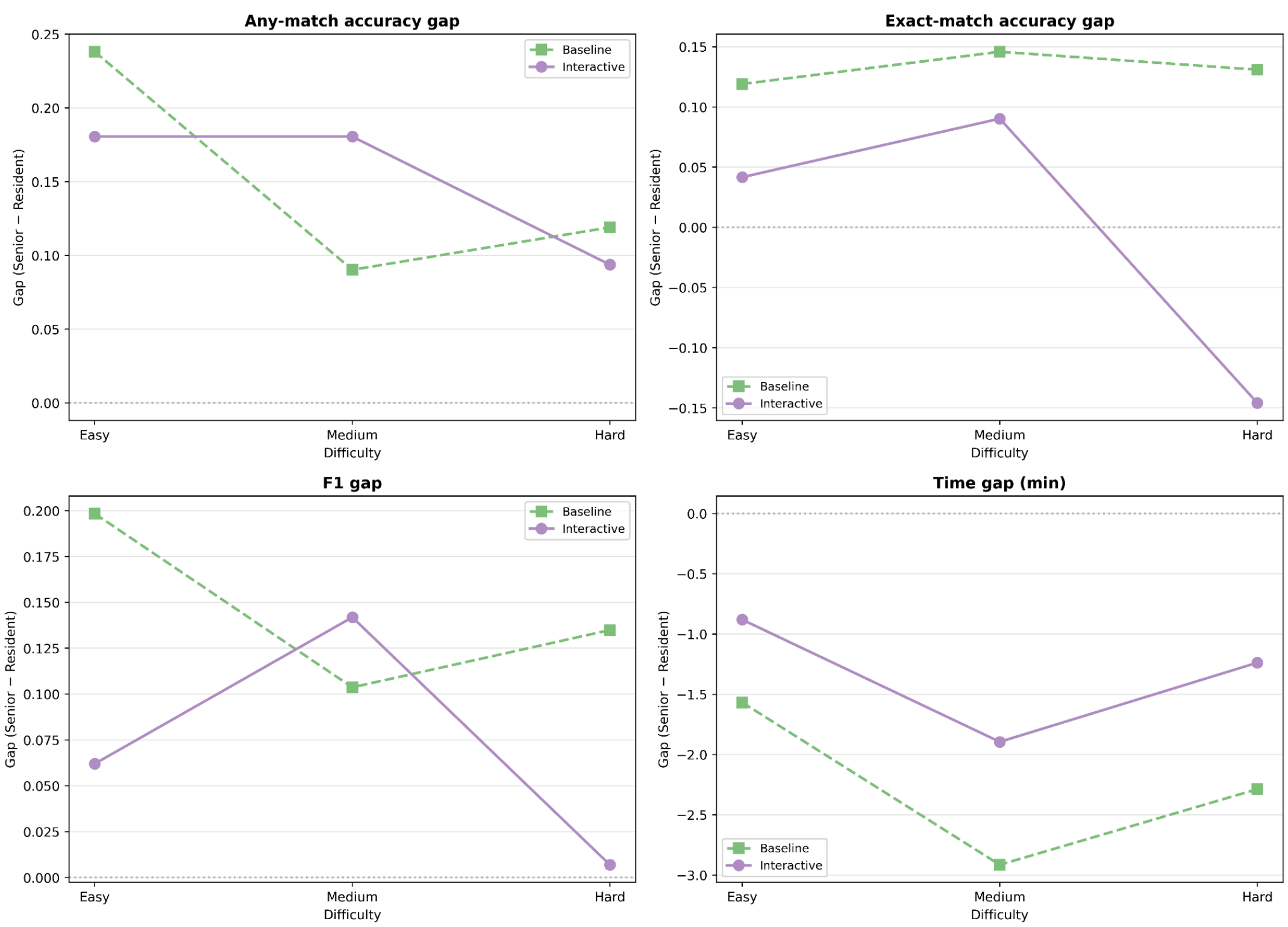

**Fig. 5 |** Four-panel line plot showing Senior-Resident gap for each metric by difficulty. Dashed: baseline, solid:interactive. The gap narrows with interactive support; on Hard-case F1 the gap inverts.

## Expert Clinical Assessment

Automated fuzzy-matching metrics assess string-level accuracy and may underestimate clinical correctness when physicians use valid but differently worded diagnoses. To obtain a clinically-grounded evaluation, a senior clinician blinded to condition assignment rated each of the 364 participant-case diagnoses as completely correct, partially correct, or wrong. Binary correctness (the proportion of diagnoses rated as at least partially correct) was high across all four sessions (Table 2), ranging from 0.912 to 0.923. This ceiling effect limits the metric's sensitivity to detect condition differences and explains why binary agreement between Sessions 1 (baseline, 0.912) and 2 (interactive, 0.923) is nearly identical. The more granular ordinal score (wrong=0, partial=0.5, complete=1) showed a modest increase from Session 1 to Session 2 (0.813→0.857) and was already high in Session 3 (0.863), possibly reflecting a practice effect in the blinded evaluator's calibration. The completely-correct rate (the most demanding criterion) improved from 0.714 (S1) to 0.791 (S2), a gain consistent with the automated F1 improvement observed for the same session pair.

**Table 2 | Blinded manual evaluation by session**

| Session | Condition | Ordinal Score | 95 % CI | Binary Correct | 95 % CI | Completely Correct | 95 % CI |
|---|---|---|---|---|---|---|---|
| **S1** | Baseline | 0.813 | [0.747, 0.879] | 0.912 | [0.857, 0.967] | 0.714 | [0.626, 0.802] |
| **S2** | Interactive | 0.857 | [0.791, 0.918] | 0.923 | [0.868, 0.978] | 0.791 | [0.703, 0.868] |
| **S3** | Baseline | 0.863 | [0.797, 0.923] | 0.912 | [0.846, 0.967] | 0.813 | [0.725, 0.890] |
| **S4** | Interactive | 0.846 | [0.780, 0.907] | 0.912 | [0.857, 0.967] | 0.780 | [0.692, 0.868] |

Difficulty-stratified manual scores (Figure 6, Table 3) reveal the most clinically significant pattern in this study. Figure 6 shows boxplots of participant-level mean ordinal scores (each participant's mean across cases of a given difficulty, averaged across sessions of the same condition), stratified by difficulty, expertise, and condition. For seniors (blue boxes), scores remain tightly clustered near ceiling across all difficulty levels and both conditions. For residents on Easy and Medium cases, baseline and interactive distributions largely overlap. The critical finding emerges on Hard cases: the Resident-Baseline box (light red) is wide and centered low (mean 0.589, median 0.571, IQR 0.464–0.696), whereas the Resident-Interactive box (dark red) shifts upward (mean 0.734, median 0.750, IQR 0.641–0.844), representing a mean gain of +0.145 ($p = 0.175$). Seniors, by contrast, scored 0.905 versus 0.896 on Hard cases ($\Delta = -0.009$, $p = 0.917$), confirming that the LLM provides no additional benefit when baseline performance is already high. On Easy cases, both groups performed near ceiling (seniors: 1.000; residents: 0.893) and interactive support conferred no measurable improvement. On Medium cases, residents showed a modest gain (+0.052, $p = 0.494$) while seniors showed a slight decline (−0.042, $p = 0.454$). Although the per-cell p-values do not reach conventional significance (consistent with the limited sample size within each difficulty × expertise stratum), the effect sizes are consistent in both direction and magnitude with the automated metric findings, and the clinical interpretation is unambiguous: LLM-mediated interaction primarily benefits less experienced clinicians confronting diagnostically complex cases.

To parallel the automated analysis (Table 1), we also computed difficulty-standardised completely-correct rates per participant (collapsing wrong and partially correct into a single category) and applied the same paired bootstrap procedure (Table 3b). Residents' completely-correct rate improved from a median of 0.580 to 0.750 (mean $\Delta = +0.092$; 95% CI, −0.021 to 0.228; $p = 0.071$; $d = 0.47$), approaching conventional significance and yielding a medium effect size consistent with the automated F1 improvement (Δ = 0.138, d = 0.94). Seniors showed a slight, non-significant decline (Δ = −0.046; p = 0.296; d = −0.28), consistent with near-ceiling baseline performance (median 0.956). At the cohort level, the improvement was small and non-significant (Δ = +0.033; p = 0.495; d = 0.21). This parallel analysis confirms that the automated and manual metrics converge on the same conclusion: interactive LLM support primarily benefits less experienced clinicians, and the lack of formal significance reflects sample size constraints rather than absence of effect.

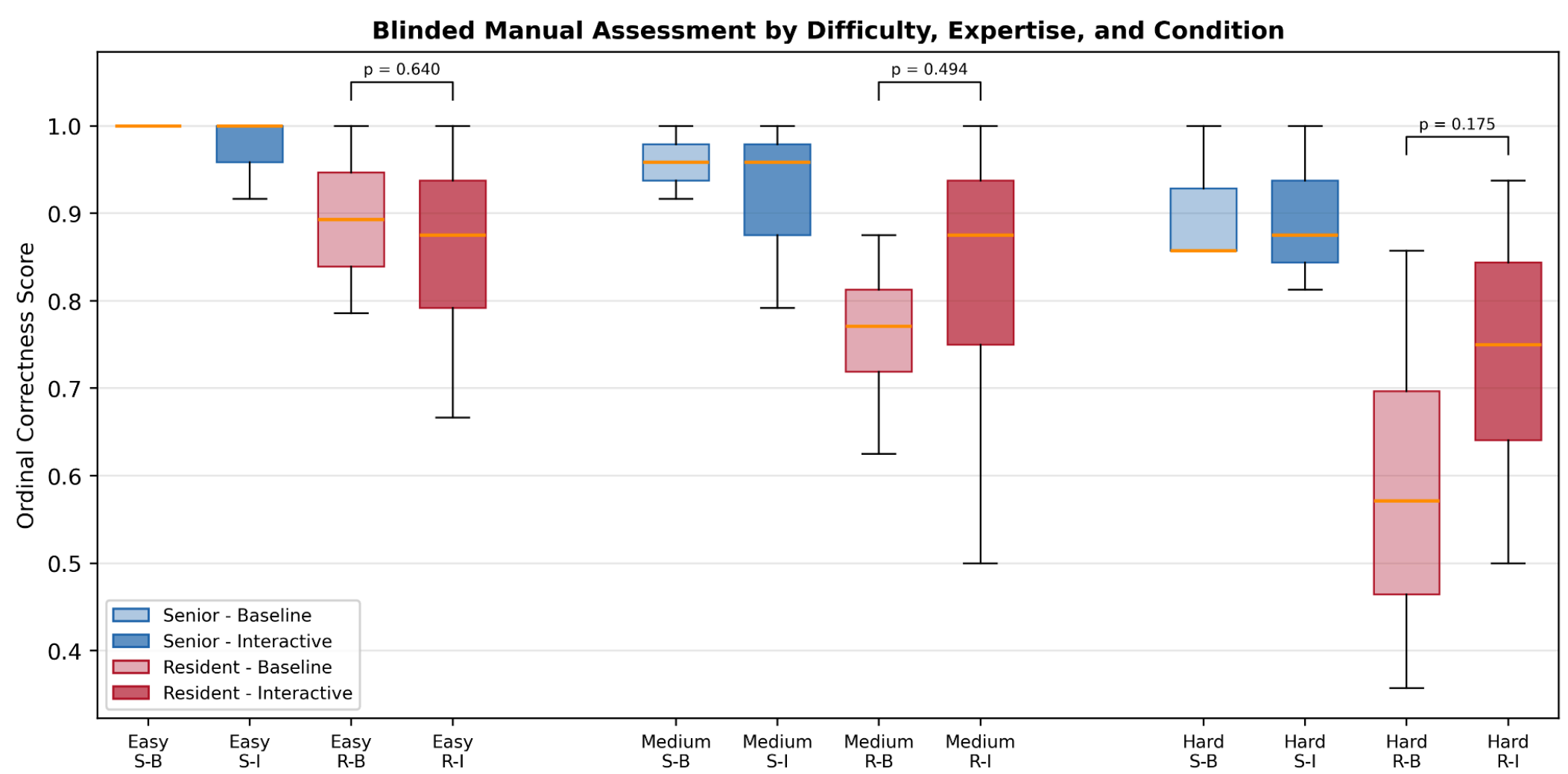

**Fig. 6 |** Boxplot of blinded manual ordinal scores (0 = wrong, 0.5 = partially correct, 1 = completely correct) by difficulty, expertise, and condition. Each box represents the distribution of per-participant mean scores (averaged across sessions of the same condition; n = 3 seniors, n = 4 residents). Blue= Senior, red=Resident; lighter shading=baseline, darker=interactive. Brackets with p-values (bootstrap, 20,000 replicates) annotate the Resident baseline–interactive comparison within each difficulty level. The largest improvement appears for residents on Hard cases (mean Δ = +0.145, p = 0.175).

**Table 3 | Blinded manual evaluation by difficulty and expertise (participant-level ordinal scores). Median [IQR] and mean of per-participant means (n = 3 seniors, n = 4 residents). p-values from case-level bootstrap (20,000 replicates).**

| Difficulty | Group | Baseline Median [IQR] | Interactive Median [IQR] | Mean Δ | p |
|---|---|---|---|---|---|
| Easy | Senior | 1.000 [1.000-1.000] | 1.000 [0.958-1.000] | −0.028 | 0.714 |
| Easy | Resident | 0.893 [0.839-0.946] | 0.875 [0.792-0.938] | −0.039 | 0.640 |
| Medium | Senior | 0.958 [0.938-0.979] | 0.958 [0.875-0.979] | −0.042 | 0.454 |
| Medium | Resident | 0.771 [0.719-0.812] | 0.875 [0.750-0.938] | +0.052 | 0.494 |
| Hard | Senior | 0.857 [0.857-0.929] | 0.875 [0.844-0.938] | −0.009 | 0.917 |
| Hard | Resident | 0.571 [0.464-0.696] | 0.750 [0.641-0.844] | +0.145 | 0.175 |

**Table 3b | Difficulty-standardised manual evaluation, completely-correct rate (bootstrap, 20,000 replicates). Completely correct = 1; partially correct and wrong = 0. Same standardisation weights (easy 3/13, medium 6/13, hard 4/13) and paired bootstrap as Table 1. † p < 0.10.**

| Group | N | Baseline Median | Interactive Median | Mean Δ | 95% CI | p-value | d |
|---|---|---|---|---|---|---|---|

| All | 7 | 0.841 | 0.885 | +0.033 | [-0.054,+0.133] | 0.495 | 0.21 |
|---|---|---|---|---|---|---|---|
| Senior | 3 | 0.956 | 0.923 | -0.046 | [-0.143,+0.044] | 0.296 | -0.28 |
| Resident | 4 | 0.580 | 0.750 | +0.092 | [-0.021,+0.228] | 0.071 † | 0.47 |

## Agreement Between Automated and Manual Evaluation

Given the discrepancy between fuzzy matching and expert judgment, we quantified their alignment at the participant-case level (N = 364). Binary agreement between any-match and manual binary correctness was 0.547, corresponding to a Cohen's kappa of 0.148, which indicates little agreement by conventional benchmarks (Table 4). The confusion matrix (Fig. 7a) shows that the disagreement was highly asymmetric: 165 of 364 cases (45.3%) were judged clinically correct by the expert reviewer but not credited by fuzzy matching, whereas no cases were credited by fuzzy matching and judged incorrect by the expert reviewer. This pattern suggests that the automated metrics were conservative, tending to under-credit clinically acceptable diagnoses rather than over-credit incorrect ones. Accordingly, the improvements observed with interactive support may be underestimated when assessed using any-match and diagnosis-set F1 alone.

**Table 4 | Automated vs manual evaluation agreement**

| Metric | Value |
|---|---|
| **Binary agreement** | 0.547 |
| **Cohen's kappa (binary)** | 0.148 |
| **Three-class raw agreement** | 0.387 |
| **True Positive (both correct)** | 168 |
| **True Negative (both wrong)** | 31 |
| **False Positive (auto ✓, manual ✗)** | 0 |
| **False Negative (auto ✗, manual ✓)** | 165 |
| **Total pairs** | 364 |

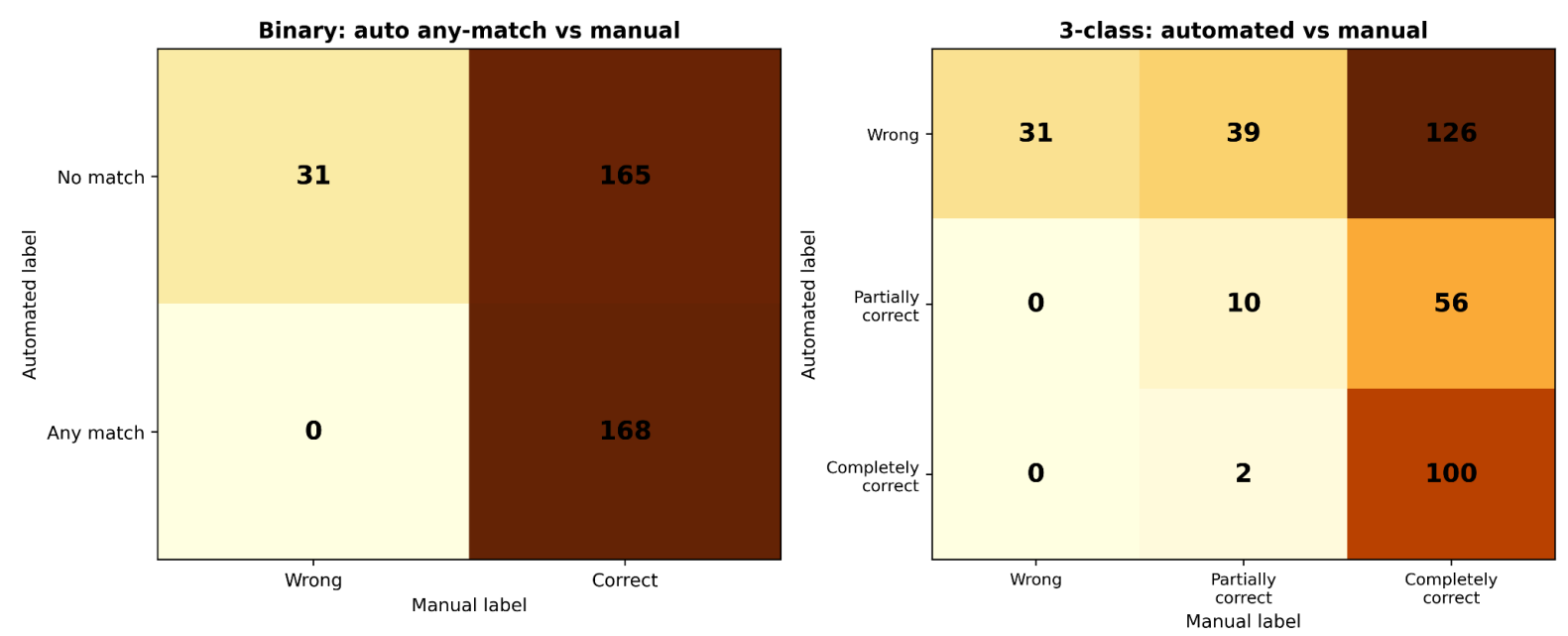

**Fig. 7 |** Binary (a) and Three-class (b) confusion matrices. The dominant off-diagonal cell (165/364) shows automated metrics under-crediting clinically correct diagnoses.

## Inter-user Concordance

The preceding analyses compare each physician's output to a ground-truth discharge diagnosis. A complementary question is whether interactive support promotes agreement among physicians themselves, independent of correctness. For each case and session, we computed pairwise diagnosis-set F1 between every clinician pair using the same fuzzy-matching rule, then separated comparisons into within-expertise (senior-senior, resident-resident) and cross-expertise (senior-resident) concordance (Figure 8). For each participant, pairwise F1 values were difficulty-standardised using the same weights as for primary endpoints. The resulting per-participant concordance scores were then compared between conditions using paired bootstrap (resampling participants, not pairs). All six concordance comparisons were highly significant (Table 5; all $p < 0.0001$). Within-expertise concordance improved by a mean of 0.100 (d = 1.41) for the full cohort, with residents showing the larger gain (Δ = 0.106; d = 1.27) than seniors (Δ = 0.093; d = 0.69).

Cross-expertise concordance (arguably the more clinically meaningful measure, as it captures whether senior and resident diagnoses converge) improved even more (Δ = 0.142; d = 1.48). This pattern suggests that the LLM acts as a shared information source that brings diagnostically diverse physicians toward a common, more complete differential. The effect is not simply that everyone copies the LLM's answers: if it were, we would expect near-perfect concordance, whereas the observed interactive medians (0.44 – 0.50) remain well below 1.0, indicating that physicians still exercise independent clinical judgement. Because concordance was computed using the same fuzzy-matching rule shown to have a 45% false-negative rate against expert judgment (Table 4), the absolute concordance values likely underestimate true inter-physician agreement. However, the relative improvement between baseline and interactive conditions is less affected, as the conservative measurement bias is approximately constant across conditions.

**Table 5 | Inter-user concordance: bootstrap tests**

| Type | Group | Baseline Median | Interactive Median | Mean Δ | 95 % CI | p | d |
|---|---|---|---|---|---|---|---|
| **Within** | All | 0.325 [0.282, 0.382] | 0.440 [0.427, 0.472] | +0.100 | [0.057, 0.142] | <.0001 | 1.41 |
| **Within** | Senior | 0.417 [0.382, 0.421] | 0.496 [0.472, 0.509] | +0.093 | [0.024, 0.175] | <.0001 | 0.69 |
| **Within** | Resident | 0.282 [0.258, 0.295] | 0.427 [0.366, 0.437] | +0.106 | [0.049, 0.147] | <.0001 | 1.27 |
| **Cross** | All | 0.341 [0.274, 0.350] | 0.490 [0.439, 0.501] | +0.142 | [0.087, 0.201] | <.0001 | 1.48 |
| **Cross** | Senior | 0.346 [0.294, 0.350] | 0.449 [0.439, 0.470] | +0.142 | [0.083, 0.249] | <.0001 | 0.88 |
| **Cross** | Resident | 0.323 [0.286, 0.351] | 0.501 [0.436, 0.521] | +0.142 | [0.064, 0.222] | <.0001 | 1.14 |

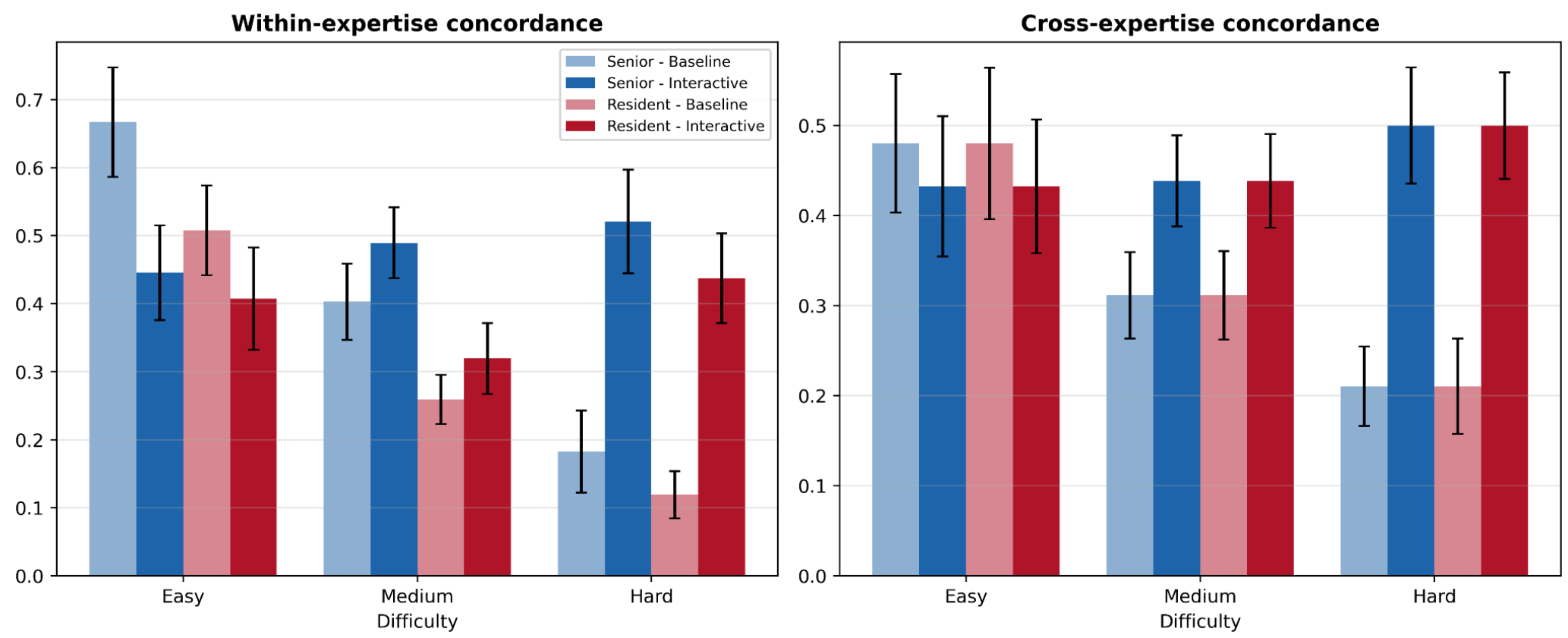

**Fig. 8 |** Within-expertise (left) and cross-expertise (right) concordance by difficulty, expertise, and condition. Concordance gains are largest on Hard cases.

## Interaction Dynamics

Across the two interactive sessions, 7 physicians generated 829 dialogue turns across 182 physician-case encounters (7 physicians × 13 cases × 2 interactive sessions; Session 2: 479 turns across 91 encounters; Session 4: 350 turns across 91 encounters). Table 6 shows that seniors used significantly more turns per case than residents (pooled: 5.14 vs 4.12; Mann-Whitney p = 0.040) yet spent comparable or less time (S2: 3.96 vs 4.26 min; S4: 3.40 vs 3.99 min). The shorter per-turn latency for seniors is consistent with a more focused, hypothesis-driven querying style that extracts targeted information efficiently. Both groups reduced their turn count between S2 and S4 (seniors: 6.23 → 4.05; residents: 4.54 → 3.69), suggesting a learning curve in which physicians become more concise in their queries as they gain familiarity with the system's capabilities.

**Table 6 | Physician-LLM interaction effort**

| Session | Group | Mean Turns | SD | Mean Duration (min) | SD | N Cases |
|---|---|---|---|---|---|---|
| **S2** | Senior | 6.23 | 3.67 | 3.96 | 5.48 | 39 |
| **S2** | Resident | 4.54 | 2.80 | 4.26 | 2.65 | 52 |
| **S4** | Senior | 4.05 | 1.83 | 3.40 | 1.38 | 39 |
| **S4** | Resident | 3.69 | 2.88 | 3.99 | 2.98 | 52 |

Question-type classification (Table 7) revealed expertise-dependent interaction strategies. Seniors devoted a larger share of their turns to information requests (S2: 44.4 % vs 29.7 %), i.e. questions that retrieve specific clinical facts such as laboratory values or imaging findings, and to detail requests that probe particular aspects of a candidate diagnosis. Residents, conversely, had a substantially higher proportion of “other” queries (S2: 41.9 % vs 25.1 %), which include broad or unstructured questions. Suggestion-seeking queries (e.g. explicitly asking the LLM for differential diagnoses) were relatively infrequent in both groups (5.5–10.8%) but increased from Session 2 to Session 4 for both, suggesting greater willingness to use the LLM for diagnostic exploration over time.

**Table 7 | Question-type distribution**

| Session | Group | Detail Req. | Info Req. | Suggestion | Other | Total |
|---|---|---|---|---|---|---|
| **S2** | Senior | 56 (23.0 %) | 108 (44.4 %) | 18 (7.4 %) | 61 (25.1 %) | 243 |
| **S2** | Resident | 54 (22.9 %) | 70 (29.7 %) | 13 (5.5 %) | 99 (41.9 %) | 236 |
| **S4** | Senior | 38 (24.1 %) | 43 (27.2 %) | 17 (10.8 %) | 60 (38.0 %) | 158 |
| **S4** | Resident | 30 (15.6 %) | 68 (35.4 %) | 19 (9.9 %) | 75 (39.1 %) | 192 |

Response quality was assessed using LLM-as-judge (google/gemini-2.5-flash[1] model) scores (faithfulness and answer relevancy) and context overlap; the fraction of answer tokens grounded in the clinical note. Table 8 summarises the statistical comparisons. Context overlap was significantly higher for senior-initiated queries (0.410 vs 0.378; $p = 0.020$), indicating that more targeted questions elicit more evidence-grounded responses. Faithfulness differed significantly by question type: information requests yielded the highest faithfulness (0.748), while suggestion-seeking queries produced the lowest (0.497; $p = 0.002$), consistent with the expectation that open-ended requests are harder to anchor in documented evidence. Answer relevancy was uniformly high (> 0.93) and did not differ by group, session, or question type, suggesting that the LLM consistently addressed the physician's intent regardless of query style.

**Table 8 | Statistical comparisons of dialogue quality (Mann-Whitney U)**

[1] https://deepmind.google/models/gemini/flash/

| Comparison | Metric | Mean A | Mean B | Diff | p |
|---|---|---|---|---|---|
| **Senior vs Resident** | Turns/case | 5.14 | 4.12 | +1.03 | .040* |
| **Senior vs Resident** | Context overlap | 0.410 | 0.378 | +0.032 | .020* |
| **Senior vs Resident** | Faithfulness | 0.642 | 0.632 | +0.010 | .938 |
| **Senior vs Resident** | Relevancy | 0.951 | 0.943 | +0.008 | .807 |
| **Session 2 vs 4** | Faithfulness | 0.644 | 0.627 | +0.017 | .683 |
| **Session 2 vs 4** | Relevancy | 0.941 | 0.953 | −0.012 | .623 |
| **Info req. vs rest** | Faithfulness | 0.748 | 0.577 | +0.172 | <.001* |
| **Suggestion vs rest** | Faithfulness | 0.497 | 0.649 | −0.152 | .002* |
| **Detail req. vs rest** | Faithfulness | 0.561 | 0.657 | −0.097 | .011* |

## Illustrative Dialogues

To illustrate how interaction quality varies with clinical expertise, we present two representative dialogues selected by a senior clinician who reviewed all interactive-session transcripts. These dialogues were chosen because they exemplify the contrasting questioning strategies identified in the quantitative analysis above: hypothesis-driven information retrieval by an expert versus LLM-guided differential reasoning by a non-expert.

***Dialogue 1: Expert physician - targeted rule-out of a Hard case***

*Senior physician, Session 2, Hard case. Chief complaint: chest pain. Diagnosis: non-cardiac chest pain (blinded assessment: partially correct). Duration: 3.6 min, 10 dialogue turns.*

| | |
|---|---|
| **Physician** | Age and sex of the patient |
| **MedSyn** | *Male; age not reported (redacted in note).* |
| **Physician** | Please give me the report of the ECG |
| **MedSyn** | *Normal sinus rhythm; nonspecific ST changes, consistent with prior ECGs.* |
| **Physician** | Troponin level? |
| **MedSyn** | *Cardiac troponin T <0.01 on admission and repeat; no elevation documented.* |
| **Physician** | D-dimer? |
| **MedSyn** | *Not reported in the note.* |
| **Physician** | Chest X-ray? |
| **MedSyn** | *No acute intrathoracic process.* |
| **Physician** | Any documentation of myocardial infarction? |
| **MedSyn** | *No. Troponins repeatedly <0.01; ECG without acute ischemic changes; chest pain brief, self-limited, non-recurrent.* |
| **Physician** | Any documentation of angina? |
| **MedSyn** | *No explicit documentation. Chief complaint is crushing substernal chest pain (~1 min, non-exertional), but angina is not diagnosed.* |
| **Physician** | Pericarditis? |
| **MedSyn** | *No. No pleuritic pain, no rub on exam, no diffuse ST elevation, troponins negative.* |
| **Physician** | Which other differential diagnoses are possible? |
| **MedSyn** | *Hypertensive urgency-related chest discomfort (BP up to 200/90); esophageal spasm/GERD (history of GERD, pain at rest during sleep); musculoskeletal pain; PE (poorly supported); anxiety-related pain (history of depression).* |
| **Physician** | Final answer: non-cardiac chest pain |

This dialogue demonstrates how a senior physician uses the LLM as an efficient information-retrieval tool within a Bayesian diagnostic framework. Following standard emergency care reasoning for chest pain, the physician first requests demographic data and key investigations (ECG, troponin, D-dimer, chest X-ray) to rapidly estimate the probability of life-threatening conditions such as acute coronary syndrome, pulmonary embolism, and pericarditis. Each query is a targeted information request designed to confirm or exclude a specific hypothesis. Only after ruling out these emergencies does the physician broaden the differential to non-life-threatening causes, using a single open-ended question. The entire interaction takes 3.6 minutes and 10 turns, consistent with the quantitative finding that seniors use more information requests (Table 7) and achieve higher context overlap (Table 8), reflecting a structured, hypothesis-driven approach.

***Dialogue 2: Resident physician - LLM-guided reasoning on a Medium case***

*Resident physician, Session 4, Medium case. Chief complaint: right back pain. Diagnosis: complicated UTI with upper-tract involvement and right-sided obstruction (blinded assessment: completely correct). Duration: 8.9 min, 5 dialogue turns.*

| | |
|---|---|
| **Physician** | Medical history and more info about the symptoms? |
| **MedSyn** | *Right back/flank pain, severe, developed over the day. Denies dysuria, fevers, chills. No history of kidney stones. Pregnant (G2P1). PMH: Chiari I malformation (s/p decompression), PTSD, anxiety, asthma, obesity. Exam: no CVA tenderness, abdomen soft and non-tender.* |
| **Physician** | Summarize imaging findings and urinary/blood exam findings relevant to the flank pain |
| **MedSyn** | *Renal US: marked asymmetric right hydronephrosis, no definite calculus but obstruction cannot be excluded. UA: RBCs 45–56/HPF, WBCs >182/HPF, leukocyte esterase large, nitrite positive. Urine culture: E. coli >100,000 CFU/mL (multi-drug resistant). Blood: Hb 8.4–10.3 g/dL (microcytic anaemia); creatinine 0.5; afebrile throughout.* |
| **Physician** | Could it be cystitis or pyelonephritis? |
| **MedSyn** | *Cystitis: consistent (pyuria, bacteriuria, no systemic signs), but insufficient to explain severe flank pain and hydronephrosis. Pyelonephritis: partially supported (flank pain, hydronephrosis, positive cultures), but no fever, no leukocytosis, no CVA tenderness.* |

| Physician | Could it be renal colic? |
|---|---|
| MedSyn | *Partially supported: severe unilateral flank pain, hematuria, hydronephrosis, and imaging cannot exclude obstruction. However, significant pyuria and positive cultures argue against a purely non-infectious stone event.* |
| Physician | What is the diagnosis in your opinion? |
| MedSyn | *Complicated UTI with upper-tract involvement and right-sided obstruction (physiologic hydronephrosis of pregnancy vs occult calculus). This integrates the UTI evidence, flank pain, hydronephrosis, and absence of systemic inflammatory response.* |
| Physician | Final answer: complicated UTI with upper-tract involvement and right-sided obstruction |

This dialogue illustrates how a non-expert physician engages in a true diagnostic dialogue with the LLM to navigate a complex clinical scenario. Rather than testing specific hypotheses, the resident begins with a broad request for medical history and symptoms, then asks the LLM to summarise only the findings relevant to the presenting complaint. The subsequent turns are suggestion-seeking: the resident proposes candidate diagnoses (cystitis, pyelonephritis, renal colic) and asks the LLM to evaluate each against the documented evidence. Crucially, the resident ultimately asks the LLM for its opinion on the most likely diagnosis; a query pattern rarely observed in senior transcripts. The LLM synthesizes the findings into an integrated diagnosis that the resident adopts. Despite taking longer (8.9 vs 3.6 minutes) and using fewer turns (5 vs 10), the resident arrives at a completely correct diagnosis as judged by the blinded evaluator.

Together, these dialogues illustrate a complementary pattern: expert physicians use the LLM primarily as an information-retrieval interface to accelerate their existing clinical reasoning, whereas non-expert physicians engage the LLM as a reasoning partner that helps structure the diagnostic process itself. Both strategies led to clinically acceptable diagnoses, but through fundamentally different interaction modes. Notably, both cases received an automated any-match score of zero despite being judged partially or completely correct by the blinded evaluator. These cases exemplify the conservative bias of string-matching metrics documented in Table 4, and underscore the importance of expert assessment in evaluating diagnostic quality.

## Clinician Perception

Post-session questionnaire (see Supplementary Note 1) comprising six five-point Likert items and 2 open-ended questions were administered after each interactive session. Internal consistency of the six Likert-scale items was good (Cronbach's[50] α = 0.81 in Session 2, 0.87 in Session 4). Diagnostic usefulness received the highest ratings across both sessions (Table 9; S2: 4.9 ± 0.4; S4: 4.6 ± 0.5), while workflow fit and willingness to recommend received the lowest (3.9 – 4.1) with the greatest variance, indicating divided opinions. Senior physicians rated MedSyn higher overall (composite 4.56 and 4.50) than residents (4.08 and 3.83); the gap was widest on workflow integration (suggesting that seniors, who are more accustomed to structured diagnostic reasoning, found the LLM's question-answer format more natural) and on willingness to recommend (reflecting residents' uncertainty about clinical adoption). Open-ended feedback highlighted three valued capabilities (synthesising long clinical notes, integrating imaging findings, and suggesting differentials for complex presentations) and three improvement areas: shorter responses, multi-language support, and explicit provenance indicators linking statements to specific note sections.

**Table 9 | Post-session Likert ratings (1-5 scale)**

| Dimension | Session 2 | Session 4 |
|---|---|---|
| **Diagnostic usefulness** | 4.9 ± 0.4 | 4.6 ± 0.5 |
| **Clarity of responses** | 4.4 ± 0.5 | 4.3 ± 0.5 |
| **Confidence in accuracy/safety** | 4.3 ± 0.8 | 4.0 ± 0.8 |
| **Time efficiency** | 4.1 ± 0.7 | 4.1 ± 0.9 |
| **Workflow fit** | 4.1 ± 1.1 | 3.9 ± 1.2 |
| **Willingness to recommend** | 3.9 ± 1.3 | 3.9 ± 1.2 |
| **Composite** | 4.28 | 4.13 |
| **Cronbach's α** | 0.81 | 0.87 |

# Discussion

This study provides controlled evidence that interactive LLM assistance improves diagnostic accuracy in emergency medicine. Automated metrics showed a mean improvement in standardised any-match accuracy of 0.156 ($p < 0.0001$; $d = 1.25$), consistent across expertise levels and reproducible across two independent interactive sessions; blinded expert assessment confirmed the direction of these gains, with residents' difficulty-standardised completely-correct rate improving by 0.092 ($p = 0.071$, $d = 0.47$). The benefit was most pronounced for residents, whose F1 improved by 0.138 ($d = 0.94$) compared to 0.075 ($d = 0.32$) for seniors, supporting the hypothesis that LLM assistance is most valuable when the physician's baseline knowledge is furthest from the diagnostic target. Critically, blinded manual assessment by a chief physician corroborated this finding: residents' mean ordinal correctness on Hard cases improved from 0.589 to 0.734 (+0.145), with the median shifting from 0.571 to 0.750, while seniors remained near ceiling (0.905 vs 0.896). These expertise-dependent gains complement recent findings from O'Sullivan et al.[13], who showed that LLM-assisted cardiologists matched or exceeded unassisted specialists in a single-specialty RCT; our results extend this evidence to emergency medicine and demonstrate that the magnitude of benefit depends on the physician's baseline expertise. This convergence of automated and clinician-judged metrics strengthens the conclusion that the observed improvements are clinically meaningful and not artifacts of string-matching sensitivity.

The difficulty-stratified analysis adds nuance to this finding. On Easy cases, interactive support conferred no benefit and may even have broadened differentials unnecessarily, as evidenced by the sharp decline in exact-match accuracy. The manual evaluation confirms this pattern: both seniors and residents scored near ceiling on Easy cases regardless of condition (seniors: 1.000 vs 0.972; residents: 0.893 vs 0.854), leaving no room for improvement. On Hard cases, by contrast, residents' any-match more than doubled and the Senior–Resident gap nearly closed. The manual scores tell an even clearer story: residents' median Hard-case correctness shifted from 0.571 to 0.750 (IQR narrowing from 0.464–0.696 to 0.641–0.844), indicating a consistent upward shift in diagnostic quality when LLM support was available. This difficulty-dependent effect has practical implications: in a deployed system, adaptive triggering (offering LLM assistance only when case complexity exceeds a threshold) could maximize benefit while avoiding the diagnostic-expansion cost on straightforward presentations.

Dialogue analysis revealed that seniors and residents interact with the LLM differently: seniors employ a hypothesis-driven strategy (more information requests, higher context overlap) while residents ask broader questions. Despite these differing strategies, cross-expertise concordance increased significantly ($\Delta$ = 0.142; d = 1.48), suggesting that the LLM functions as a shared knowledge source that brings diagnostically diverse physicians toward a common, more complete differential without homogenising their reasoning.

These findings carry several implications for the design and deployment of AI-assisted clinical workflows. First, the difficulty-dependent benefit profile suggests that adaptive deployment strategies, in which LLM assistance is offered selectively for cases exceeding a complexity threshold, could maximize diagnostic gains while avoiding the unnecessary broadening of differentials observed on straightforward presentations. Second, the expertise-stratified interaction patterns point to a need for differentiated interface design: senior physicians, who use the LLM primarily for targeted information retrieval, may benefit from concise, evidence-linked responses, whereas residents, who engage in more open-ended diagnostic reasoning, may benefit from structured scaffolding that guides them through differential diagnosis. Third, the finding that residents achieve the greatest gains in accuracy has implications for medical education: LLM-based tools could serve as a training resource that exposes trainees to structured clinical reasoning on complex cases, complementing traditional supervision. Finally, the significant increase in cross-expertise concordance suggests that LLM-mediated interaction may help standardize diagnostic quality across experience levels, a particularly relevant consideration for settings with heterogeneous staffing, such as emergency departments during off-peak hours.

The modest agreement between automated and manual evaluation ($\kappa$ = 0.148) is an important methodological finding. The false-negative rate of automated metrics (45% of cases) means that the improvements reported here are likely conservative. The difficulty-stratified manual scores (Table 3) provide a more clinically grounded view: the +0.145 improvement in residents' Hard-case ordinal scores and the shift from median 0.571 to 0.750 are arguably the most compelling evidence that interactive LLM support has tangible clinical value, particularly for less experienced physicians facing complex diagnostic challenges. Future work should develop clinically-validated matching algorithms (possibly using medical ontologies rather than string similarity) to narrow this gap.

Several limitations qualify these results. The sample (7 physicians, 52 cases) limits generalisability; statistically significant effects should be interpreted as directional evidence. When the same difficulty-standardised bootstrap procedure used for automated endpoints (Table 1) is applied to the manual completely-correct rate, the Resident improvement approaches but does not reach conventional significance ($\Delta$ = +0.092, p = 0.071, d = 0.47). This reflects the near-ceiling baseline scores (manual baseline mean 0.83 vs. automated 0.37), the coarse three-point ordinal scale, and the small sample (n = 7), which together limit the sensitivity of the manual metric. Three of seven participants showed slight manual score decreases despite automated metric improvements, indicating that string-level concordance and clinical correctness capture different facets of diagnostic quality. We therefore interpret the manual evaluation as a clinically grounded complement to the automated analysis: it confirms effect direction, localises the benefit to residents on Hard cases (Table 3), and provides the most compelling evidence through difficulty-stratified effect sizes rather than aggregate significance tests. The within-subjects design introduces potential learning effects, although the S3 baseline reversion argues against persistent carryover. The use of retrospective MIMIC-IV notes rather than real-time encounters omits time pressure, interruptions, and evolving clinical information. Additionally, manual evaluation was performed by a single blinded rater; inter-rater reliability was not assessed, and individual calibration effects cannot be excluded.

Future work should evaluate MedSyn in prospective clinical settings with larger cohorts. The framework is not limited to emergency medicine; adaptation to internal medicine, where documentation burden and diagnostic complexity are similarly high, is a natural extension. Priorities identified by participants include shorter responses, multilingual support, and explicit provenance linking answers to specific note sections.

# Methods

### Dataset

We constructed our study dataset by combining MIMIC-IV[2] and MIMIC-IV-Note[3] records and selecting samples with ICD-10 diagnosis codes[4]. ICD-10 codes span multiple levels of granularity, from coarse "chapter-level" ranges (e.g., E00-E90) to fine-grained codes (e.g., E10.9, where E10 denotes the disease category and .9 specifies a subtype). The initial merged dataset contained 122,266 hospitalization records and 5,802 unique diagnoses. Because our experiments rely on extracting clinically meaningful context from discharge documentation, we examined the discharge text field and found that while many notes follow a semi-standard structure, subsection presence and formatting vary substantially (e.g., "major surgical or invasive procedure" may be present or absent). Notes with missing headings or highly free-form text impeded reliable parsing and prevented enforcing a consistent input format across cases. After consultation with three physicians, we identified the most informative discharge note sections for diagnostic reasoning and retained only records whose discharge texts contained the following headings: chief complaint; history of present illness; social history; physical exam; pertinent results; major surgical or invasive procedure; brief hospital course; medications on admission; discharge medications; discharge diagnosis; discharge condition; and discharge instructions. We additionally excluded records where the patient status was "deceased" or "expired". This filtering yielded a final dataset of 74,850 records. From this filtered subset, we randomly sampled 1,000 records[5] as the test set (seed = 13). The resulting test set contains 2,350 unique diagnoses (13,384 total code occurrences), with an average of 5.61 ICD-10 codes per record.

[2] https://physionet.org/content/mimiciv/3.0/
[3] https://www.physionet.org/content/mimic-iv-note/2.2/
[4] https://icd.who.int/browse10/2019/en (Accessed on 2026-01-27)
[5] One record lacked a ground-truth diagnosis and was removed from evaluation.

**Data access and reproducibility:** Access to MIMIC requires completion of the PhysioNet credentialing process, including CITI training[6]. Accordingly, we cannot publicly release the selected test set. To support reproducibility, we provide the complete preprocessing and evaluation pipeline and release our code[7].

## Use cases

We consider two use cases: (i) *baseline* (unassisted diagnosis), in which physicians review the complete clinical record and provide a diagnosis without LLM support; and (ii) *interactive* (AI-supported diagnosis), in which physicians initially view only the chief complaint and iteratively query an LLM (granted access to the full record) to retrieve additional information needed before finalizing their diagnosis.

## Choice of LLM for Experiments

We evaluated MedSyn in two phases. Phase 1 consisted of a controlled, multi-agent simulation in which an LLM "chief physician" interacted with an LLM "assistant physician." The chief physician was limited to the patient's chief complaint, while the assistant had access to the full clinical note; the dialogue continued until the chief physician finalized a diagnosis-focused discharge output. To enable automated scoring under controlled conditions, we used ICD-10 codes as the diagnosis endpoint in this phase. Overall, Phase 1 was intended to stress-test multi-turn interaction dynamics, compare model behaviors, and characterize diagnostic overlap across models.

In brief, these preliminary simulations suggested that role-separated dialogue can improve diagnostic completeness: complaint-constrained reasoning by a chief physician coupled with on-demand note extraction by an assistant can yield more clinically coherent discharge narratives and may help surface less common conditions. At the same time, performance varied substantially across models and was strongly influenced by instruction adherence and the ability to ask targeted follow-up questions. The simulation results also highlighted an important evaluation consideration: reference ICD-10 lists often contain chronic or historical comorbidities recorded through coding workflows, whereas MedSyn prompts emphasize the current presentation, making exact code-level comparisons potentially misleading. Coarser, set-based evaluation at the chapter/category level (e.g., precision/recall/F1/Jaccard) was therefore more interpretable, and under this lens models were generally better at identifying the broad disease domain than the specific ICD category. Full details of Phase 1 are reported in our workshop paper[48].

Phase 2 replaced the simulated chief physician with real physicians and evaluated MedSyn as an interactive decision-support workflow. Physicians initially saw only the chief complaint and iteratively queried the assistant (with access to the full record) before committing to a final diagnosis. In this phase, physicians reported diagnoses in free text, reflecting real clinical practice, and we compared their outputs against ground-truth textual discharge diagnoses. This paper focuses on the Phase 2 study design, outcomes, and clinician-facing findings.

---

[6] https://physionet.org/about/citi-course/ (Accessed on 2026-01-27)
[7] https://github.com/burcusayin/MedSyn

To select the LLM for physician-facing (real-world) experiments, we evaluated five recent models for the time of this study (gemini-3-pro-preview[8], gpt-oss-120b[51], llama-4-scout[9], gpt-5.1-chat, and gpt-5.2-chat[52]) on 52 cases under both baseline and interactive scenarios, scoring against primary discharge diagnoses using fuzzy matching (RapidFuzz[10] token-set ratio, threshold 62). In the baseline scenario, gpt-oss-120b achieved the highest micro-F1 (0.626), followed by gpt-5.2-chat (0.598) and gemini-3-pro-preview (0.590). In the interactive scenario, llama-4-scout led in micro-F1 (0.576), with gpt-5.2-chat close behind (0.562). These ablation results measure standalone model performance without physician involvement and were used solely for model selection; they are not directly comparable to the physician-facing results in Table 1, where physicians actively reason over the model's outputs. Based on the results, we selected gpt-5.2-chat for deployment because it offered the best trade-off between F1, recall (0.725), top-1 accuracy (0.808), and output length (2.1 diagnoses per case), balancing diagnostic coverage with manageable cognitive load for clinicians. Full ablation results, including difficulty-stratified analyses and bootstrap confidence intervals, are reported in Supplementary Table 1 and 2.

### Stakeholders

We evaluated our framework with seven physicians: three internal medicine experts (senior physicians) and four first-year residents (non-expert physicians). An additional senior physician, who did not participate in the diagnostic sessions, served as the blinded evaluator for manual assessment. All participants completed the required CITI training for access to MIMIC and provided written informed consent.

### Sample Selection

The selection of clinical cases was performed through evaluation of the MIMIC database. Clinical records were reviewed sequentially according to the following exclusion criteria:

- the presenting symptom directly corresponded to the final diagnosis (e.g., gunshot wound);
- the medical diagnosis itself was recorded in the clinical chart as the presenting symptom (e.g., presenting symptom: pulmonary embolism);
- elective hospital admission for a planned procedure or therapeutic intervention;
- incomplete medical records in which instrumental and laboratory investigations, at the discretion of the evaluating physician, were insufficient to establish a definitive diagnosis.

Once selected, cases were stratified according to their level of difficulty (easy, intermediate, and difficult), as determined by the evaluating physician. In particular, the most challenging cases corresponded to diagnoses of rare or less common diseases.

### Sessions

The study included four sessions. Each session comprised 13 cases. Because Session 3 differed slightly in case-difficulty composition from the other sessions (4 easy and 3 hard cases, versus 3 easy and 4 hard cases), all between-session comparisons were difficulty-standardized using fixed weights for easy, medium, and hard cases (3/13, 6/13, and 4/13, respectively). Session order was fixed (baseline–interactive–baseline–interactive) rather than randomised, because counterbalancing would have required withholding the LLM in later sessions from physicians who had already experienced it, raising ethical and motivational concerns. The S3 baseline reversion (performance returning to near-S1 levels) provides empirical evidence against persistent carryover effects.

[8] https://deepmind.google/models/gemini/
[9] https://ai.meta.com/blog/llama-4-multimodal-intelligence/
[10] https://rapidfuzz.github.io/RapidFuzz/

## Evaluation

Our success criteria are twofold: (i) reducing the time and effort required for clinical decision-making, and (ii) improving the quality of the final decision. Within the MedSyn framework, we define the final decision as the diagnosis names rather than ICD codes, as the latter are primarily used for administrative purposes, such as assessing patient complexity or determining discharge costs, and lack the clinical granularity necessary for evaluating LLM-based diagnostic performance. Relying on diagnosis names, however, presents challenges for automated evaluation. Ground truth diagnoses often consist of comprehensive or detailed lists (e.g., "heart failure" and "congestive heart failure"), whereas clinician-generated outputs may include shorter, yet clinically equivalent, formulations. To address this discrepancy, we involve a senior clinician as a blinded evaluator in our user experiments. Acting as an independent referee, the clinician manually compares the generated diagnoses with the ground truth, assessing both accuracy and clinical concordance. Dialogue quality was additionally assessed using an LLM-as-judge approach (google/gemini-2.5-flash, chosen to avoid self-evaluation bias, as the assistant uses a GPT model) that scored each response for faithfulness and answer relevancy.

## Metrics

For each participant-case pair, clinician-entered diagnoses were compared with the reference discharge diagnosis. Because both could contain multiple diagnoses, we treated each as a diagnosis set and evaluated concordance after text normalization and fuzzy string matching (RapidFuzz token-set ratio; a predicted diagnosis was considered a match if the similarity score reached or exceeded 80 on a 0-100 scale) to account for minor phrasing differences. Predicted and reference diagnoses were matched one-to-one using the highest similarity scores, so that each predicted diagnosis could match at most one reference diagnosis and vice versa. Based on the number of matched diagnosis pairs, we calculated per-case precision, recall, and F1 score to quantify partial diagnostic concordance. Precision reflected the proportion of clinician-proposed diagnoses that matched the reference set, whereas recall reflected the proportion of reference diagnoses recovered by the clinician. F1 summarized the balance between these two quantities. We additionally computed any-match accuracy, defined as whether at least one diagnosis matched the reference set, and exact-match accuracy, defined as complete agreement between clinician and reference diagnosis sets under the same matching procedure. Time per case was measured in minutes. Because sessions differed slightly in their distribution of case difficulty, all summary metrics were difficulty-standardized to a fixed target composition of easy, medium, and hard cases (3/13, 6/13, and 4/13, respectively). For each participant and session, metric values were first averaged within each difficulty level and then combined using these prespecified weights. This yielded standardized performance estimates that enabled fair comparison between baseline and interactive conditions.

**Statistical Analysis.** All between-condition comparisons used paired bootstrap resampling (20,000 replicates, seed 42). For each bootstrap replicate, participants were resampled with replacement and the mean baseline–interactive difference was computed; 95% confidence intervals were derived from the 2.5th and 97.5th percentiles of the bootstrap distribution, and two-sided p-values were computed as twice the smaller tail probability. Effect sizes are reported as Hedges' g, a small-sample bias-corrected variant of Cohen's d (correction factor $J = 1 - 3/(4n - 5)$). The primary automated outcome measure was difficulty-standardised any-match accuracy; all other automated metrics (exact-match, F1, precision, recall, time), manual evaluation scores, and concordance comparisons are reported as secondary or exploratory endpoints. No correction for multiple comparisons was applied; significant results on secondary endpoints should be interpreted as hypothesis-generating. The significance threshold was set at $\alpha = 0.05$.

## Pilot Studies & Training for Physicians

In a pilot study conducted on December 11, 2025, participants completed two sequential sessions (baseline and interactive) in a single sitting. All cases were selected to have Medium difficulty. Two cases were assigned to each session: the baseline session used 10259153-DS-2 and 12583650-DS-13, while the interactive session used 11880923-DS-18 and 18101466-DS-4.

The pilot study was a 90-minute training block. During the first 30 minutes, we introduced the study task and guided participants through the MedSyn interface and workflow; the remaining training time was used to answer questions and ensure familiarity with the interaction format. Participants then completed the baseline session first (20 minutes), followed by a 10-minute Q&A/feedback period. Next, they completed the interactive session (20 minutes), again followed by a 10-minute Q&A/feedback period. This protocol was designed to (i) standardize participants' understanding of the task and interface prior to data collection, (ii) support within-participant comparison between baseline and interactive conditions, and (iii) elicit qualitative feedback immediately after each session while the experience was fresh.

## Cost for user experiments

In the user experiments, conducted between December 15, 2025 and January 6, 2026, we used the openai/gpt-5.2-chat-20251211 model via OpenRouter API (temperature = 0.7, default parameters) and spent $7.38, spread across the study window (1,209 generations; ~3.76M tokens) with a mean cost of $0.00611 per generation, i.e., well below one cent per interaction on average.

## Ethics approval and consent to participate

The study protocol was reviewed by the Research Ethics Committee of the University of Trento through the institutional Ethics Self-Assessment procedure. The Committee confirmed with identification number 2024-088ESA that this work does not entail risks for the psychophysical well-being of the involved subjects that could potentially limit their rights to privacy, information, and decisional autonomy. All participants provided written informed consent prior to enrolment. The consent form stated that participation was voluntary, that participants could withdraw at any time without giving a reason, and that all data would be processed anonymously in accordance with the aims of the study. The study was conducted between 15 December 2025 and 6 January 2026 at the University of Trento.

# Data Availability

Clinical data were obtained from MIMIC-IV (v3.0) and MIMIC-IV-Note (v2.2), available through PhysioNet upon completion of the credentialing process. From these sources, a senior physician selected and curated 52 emergency-medicine cases stratified by difficulty (13 easy, 26 medium, 13 hard), excluding cases with self-evident diagnoses, incomplete records, or elective admissions (see Methods: Sample Selection). The curated dataset, including chief complaints, clinical notes with ground-truth diagnoses removed, difficulty labels, and reference discharge diagnoses, will be released on PhysioNet upon acceptance of the paper.

# Code Availability

The MedSyn source code, evaluation pipeline, and experiment configuration are publicly available at [https://github.com/burcusayin/MedSyn-clinical-chatbot](https://github.com/burcusayin/MedSyn-clinical-chatbot).

# Acknowledgements

Funded by the European Union. Views and opinions expressed are however those of the author(s) only and do not necessarily reflect those of the European Union or the European Health and Digital Executive Agency (HaDEA). Neither the European Union nor the granting authority can be held responsible for them. Grant Agreement no. 101120763 - TANGO.

# Author Contributions

B.S. and A.P. conceived and designed the study. A.P. supervised the project. J.S. and P.M. provided feedback on the study design and experiments. B.S. and N.V.H. devised and prepared the dataset. B.S. implemented the MedSyn framework, conducted the experiments, analysed the results, and drafted the manuscript. I.B.S. contributed to the analysis of the experimental results. N.V.H., N.O., N.S., and A.M. provided feedback on the design and evaluation of the physician-facing experiments. S.A. contributed to the preliminary design of the MedSyn framework, planning of ethical clearance, and manuscript drafting. V.R. provided strategic guidance and contributed to the manual evaluation of the experimental results. All authors proofread the manuscript and provided feedback.

# Competing Interests

The authors declare no competing interests.

# Supplementary Material

# Human–LLM Dialogue Improves Diagnostic Accuracy in Emergency Care

## Ablation results for model selection in user experiments

To better mirror real-world clinical practice, we evaluated the quality of discharge diagnosis predictions across both the baseline setting and the interactive MedSyn scenarios. To identify the model most suitable for collaborative use with physicians, we ran ablation experiments with five strong, recent LLMs: gemini-3-pro-preview, gpt-oss-120b, llama-4-scout, gpt-5.1-chat, and gpt-5.2-chat. Prompts used for ablation tests are provided in Supplementary Note 3.

The goal was to evaluate how well each model predicts discharge diagnoses under: (i) Baseline scenario: model sees the full note and outputs diagnoses directly, and (ii) Interactive scenario: model outputs diagnoses after the interactive workflow. The key decision for deployment was: which model will perform best in real physician-facing interactive experiments. Each scenario consists of 52 cases. Since the ground truth diagnosis strings and the expected physician answers are not guaranteed to be identical (abbreviations, word order), we used soft matching to evaluate models' answers: RapidFuzz[11] token set ratio on normalized strings. A prediction matches a label if similarity >= 62 (0–100 scale). When multiple predictions could match multiple labels, we perform one-to-one matching (each pred and label can be used at most once) by selecting the highest-similarity pairs first.

We report results at the diagnosis level, with a focus on primary discharge diagnoses. For each model and scenario, we computed: (i) micro-precision (P): fraction of predicted diagnoses that match a ground-truth primary diagnosis, (ii) micro-recall (R): fraction of ground-truth primary diagnoses recovered by any prediction, (iii) micro-F1: harmonic mean of (P, R), (iv) Top-1 accuracy: whether the *first predicted diagnosis* matches any primary diagnosis, (v) Any-match accuracy: whether *any predicted diagnosis* matches any primary diagnosis for that case, and (vi) Average number of predicted diagnoses (avg_n_pred): proxy for verbosity / cognitive load.

[11] https://github.com/rapidfuzz/RapidFuzz

**Supplementary Table 1. Results on Baseline Scenario:**

| Model | micro_P | micro_R | micro_F1 | top1_acc | any-match acc | avg_n_pred |
|---|---|---|---|---|---|---|
| gpt-oss-120b | **0.543** | 0.739 | **0.626** | 0.846 | 0.885 | **1.808** |
| gpt-5.2-chat | 0.495 | 0.754 | 0.598 | **0.865** | 0.923 | 2.019 |
| gemini-3-pro-preview | 0.490 | 0.739 | 0.590 | 0.788 | 0.865 | 2.000 |
| llama-4-scout | 0.257 | **0.841** | 0.393 | 0.750 | **0.942** | 4.346 |
| gpt-5.1-chat | 0.161 | 0.812 | 0.269 | 0.827 | 0.904 | 6.673 |

All models achieve relatively high recall on primary diagnoses (R ≈ 0.74-0.84), but differ markedly in precision and verbosity. *gpt-oss-120b* attains the highest micro-F1 (0.63), with a compact output (≈1.8 diagnoses per case). *gpt-5.2-chat* and *gemini-3-pro-preview* are very close in F1 (0.60 and 0.59), with similar recall (~0.74-0.75) and slightly higher average list length (~2 diagnoses). *llama-4-scout* and *gpt-5.1-chat* achieve very high recall (≥0.81) but at the cost of very low precision and long lists (4-7 diagnoses per case), which would be burdensome in a physician UI. Overall, in the baseline setting, *gpt-oss-120b*, *gpt-5.2-chat*, and *gemini-3-pro-preview* form the top tier in terms of F1, with *gpt-5.2-chat* offering a good balance between recall and output length.

**Supplementary Table 2. Results on Interactive Scenario:**

| Model | micro_P | micro_R | micro_F1 | top1_acc | any-match acc | avg_n_pred |
|---|---|---|---|---|---|---|
| gpt-oss-120b | 0.414 | 0.667 | 0.511 | 0.769 | 0.769 | 2.135 |
| gpt-5.2-chat | 0.459 | 0.725 | 0.562 | 0.808 | 0.846 | 2.096 |
| gemini-3-pro-preview | 0.354 | 0.739 | 0.479 | 0.750 | 0.865 | 2.769 |
| llama-4-scout | **0.485** | 0.710 | **0.576** | 0.712 | 0.846 | **1.942** |
| gpt-5.1-chat | 0.403 | **0.783** | 0.532 | **0.885** | **0.923** | 2.577 |

In the interactive scenario, *llama-4-scout* achieves the highest micro-F1 (0.58), with reasonably short lists (~1.9 diagnoses per case). *gpt-5.2-chat* is a close second in F1 (0.56), with slightly higher recall (0.73) but somewhat lower precision. *gpt-5.1-chat* is particularly strong on recall (0.78) and top-1 accuracy (0.89), while keeping the list length moderate (~2.6 diagnoses). This makes it attractive when the UI emphasizes the first suggestion and when missing the primary diagnosis is considered costly. *gemini-3-pro-preview* and *gpt-oss-120b* perform reasonably well but are not dominant under this metric.

**Cost of ablation experiments.** In the ablation experiments, conducted on December 13-14, 2025, we tested gemini-3-pro-preview, gpt-oss-120b, gpt-5.1-chat, gpt-5.2-chat and llama-4-scout via

OpenRouter API. We spent $13.85 in total, (1,548 generations; ~6.64M tokens) with a mean cost of $0.00895 per generation.

## Supplementary Note 1: Questionnaire used in qualitative analysis

This short survey aims to collect your feedback on MedSyn, an AI-assisted clinical decision support tool.
Your responses will help us understand how useful and trustworthy MedSyn is in realistic clinical scenarios, and guide future improvements.
The survey is completely anonymous: we do not collect names, email addresses, or any other personal identifiers.

The questionnaire takes about 5–10 minutes to complete. Participation is entirely voluntary, and you may stop at any time by closing this page.

By proceeding, you confirm that you have read this information and agree to participate in this anonymous evaluation. We sincerely thank you for your time and valuable feedback.

**1.** MedSyn provides information that is useful and sufficiently detailed to support my diagnostic decisions. (1 (Strongly Disagree) to 5 (Strongly Agree))
    1.1. Could you briefly explain your answer to Question 1? (optional)
**2.** In which situations is MedSyn particularly helpful or not helpful for diagnosis?
**3.** MedSyn's answers to my questions are clear and easy to understand. (1 (Strongly Disagree) to 5 (Strongly Agree))
    3.1. Could you briefly explain your answer to Question 3? (optional)
**4.** I feel confident that MedSyn's outputs are generally accurate and safe to use as a decision-support tool. (1 (Strongly Disagree) to 5 (Strongly Agree))
    4.1. Could you briefly explain your answer to Question 4? (optional)
**5.** How could MedSyn be improved so that you would feel more comfortable using it and relying on it in your daily work?
**6.** Using MedSyn saves me time when working up a case. (1 (Strongly Disagree) to 5 (Strongly Agree))
    6.1. Could you briefly explain your answer to Question 6? (optional)
**7.** I could imagine using MedSyn regularly as part of my clinical workflow. (1 (Strongly Disagree) to 5 (Strongly Agree))
    7.1. Could you briefly explain your answer to Question 7? (optional)
**8.** I would recommend MedSyn to colleagues. (1 (Strongly Disagree) to 5 (Strongly Agree))
    8.1. Could you briefly explain your answer to Question 8? (optional)

# Supplementary Note 2: Prompts used in experiments with simulated physician interactions

**Prompt for the assistant LLM (MedSyn agent)**

You are Dr. Lee, an assistant physician working under the supervision of Dr. Ellis, the chief physician. Your role is to review the patient’s clinical notes to perform an initial evaluation, which will support Dr. Ellis in assessing the patient’s condition and writing the discharge text. Following your evaluation, you will engage in a collaborative discussion with Dr. Ellis to confirm the diagnosis and determine the next steps.

**Here is the clinical note for the patient:** {clinicalNote}.

### Task: Thoroughly analyze the clinical note and provide a structured summary that includes:
• Key symptoms: Highlight notable symptoms that may require further investigation.
• Preliminary diagnosis: Offer an initial diagnosis based on the patient’s symptoms and medical history.
• Potential complications: Identify any critical issues or risks Dr. Ellis should consider.
• Recommendations: Suggest further evaluations if uncertainties remain about the patient’s condition.

### Instructions:
1. Ensure your evaluation is clear, precise, and structured to facilitate an informed discussion.
2. In each round of the discussion, limit yourself to a CONCISE message.
3. Keep in mind that you have limited time for every patient. Please avoid duplicate recommendations,
conversations, and questions related to treatments.

### Process: You will first receive a message from Dr. Ellis, asking for your initial assessment. Afterward, you can follow up in each discussion round to collaboratively refine the diagnosis.

**Prompt for the physician LLM**

Baseline case:

You are Dr. Ellis, the chief physician responsible for reviewing clinical notes and writing a discharge text for patients.

**Here is the clinical note for the patient:** {clinicalNote}.

### Instructions:
1. Carefully analyze the given clinical note to ensure that no symptoms are overlooked.
2. You are not allowed to ask any questions or make assumptions beyond the information provided

*in the clinical note.*
*3. Once you are ready, write the discharge text for the patient.*
*4. The discharge text should include only the 'diagnosis' and 'codes' fields:*
*• 'diagnosis' field should specify the patient's final diagnosis. Please note that you should decide the final diagnosis.*
*• 'codes' field should list the ICD-10 codes corresponding to the diagnosis specified in the 'diagnosis' field. Keep in mind that this field is a string, do not use '[]' while listing the codes.*
*5. Remember to refer the clinical note while writing the discharge text. Ensure that the 'diagnosis', and 'codes' fields are complete and unambiguous; they must not be left empty or unclear.*
*6. Return your dischargeText using the TOOL 'baseline_discharge_text_tool'.*

*Interactive case:*

*You are Dr. Ellis, the Chief Physician, collaborating with Dr. Lee, your assistant. Your task is to review a clinical note by initiating an evaluation from Dr. Lee and engaging in a natural, focused conversation to assess the patient's condition. Avoid fabricating interactions or simulating dialogue with Dr. Lee. Instead, clearly articulate your questions or follow-ups, analyze Dr. Lee's responses, and use this information to guide your decision-making. Your responsibilities include the following:*
*• Verify the patient's condition, symptoms, and diagnosis.*
*• Ensure all symptoms are accounted for and thoroughly understand the patient's condition to deliver optimal care.*
*• Address doubts regarding the diagnosis or treatment plan by conducting further evaluations with Dr. Lee to achieve accurate and effective results.*
***Here is the clinical note for the patient:** {clinicalNote}.*

*### Instructions:*
*1. Begin by requesting an initial evaluation of the patient from Dr. Lee.*
*2. Engage in a collaborative discussion to confirm the patient's diagnosis. Please note that Dr. Lee has access to a more detailed clinical note, so you MUST consult to Dr. Lee to obtain the necessary information for making the diagnosis.*
*3. Keep in mind that you have limited time for every patient. Please avoid duplicate recommendations,*
*conversations, and questions related to treatments. Keep each message CONCISE and to the point.*
*4. Once you have gathered sufficient information and are confident in the diagnosis, stop the discussion and write the patient's discharge text.*
*5. The discharge text should include only the 'diagnosis' and 'codes' fields:*
*• 'diagnosis' field should specify the patient's final diagnosis. Please note that you should decide the final diagnosis.*
*• 'codes' field should list the ICD-10 codes corresponding to the diagnosis specified in the 'diagnosis' field.*
*6. Remember to refer to your discussion with Dr. Lee and the clinical note while writing the discharge text. Ensure that the 'diagnosis', and 'codes' fields are complete and unambiguous; they must not be left empty or unclear.*

7. Do NOT ask Dr. Lee to check or write your dischargeText. It is YOUR RESPONSIBILITY to write and submit the dischargeText.
8. Return your dischargeText using the TOOL ‘discharge_text_tool’. Do NOT mention the TOOL ‘discharge_text_tool’ to Dr. Lee.

## Supplementary Note 3: Prompts used in ablation studies

### Prompt for the assistant LLM (MedSyn agent)

You are MedSyn, an AI physician assistant working under the supervision of the Chief Physician. Your role is to review the patient’s clinical notes to perform an initial evaluation, which will support the chief physician in assessing the patient's condition and writing the discharge text. Following your evaluation, you will engage in a collaborative discussion with the chief physician to confirm the diagnosis. You always see the full clinical note. You never say that you don’t have access to it.

**Here is the clinical note for the current patient:** '''{clinicalNote}'''.

----------------
ROLE & TASK
----------------

Your role is to:
- Analyze the clinical note.
- Provide clear, structured clinical information.
- Help the Chief Physician arrive at a final diagnosis and discharge text.

When the Chief Physician sends you a message, you respond **only as MedSyn**.
You never write what the Chief Physician says.

----------------
WHAT TO PROVIDE
----------------

1. **Initial evaluation (when asked for it)**
   When the Chief Physician asks for an initial evaluation, respond with a concise structured summary including:

   - **Key symptoms**: Important complaints, exam findings, and relevant history.
   - **Preliminary diagnosis**: The most likely diagnosis or a short prioritized differential.

2. **Follow-up questions from the Chief Physician**
   - Answer **only** the specific questions asked.

*- Use **only** information present in the clinical note.*
*- If the note truly does not contain the requested detail, say so clearly (e.g., “The note does not specify the patient’s medication list.”).*

*----------------*
*CONSTRAINTS*
*----------------*

*3. **You must NOT:***
*- Ask the Chief Physician to order new tests, imaging, or consultations.*
*- Ask for external data beyond the clinical note.*
*- Talk about tools, functions, prompts, or policies.*
*- Role-play or write messages as the Chief Physician.*
*- Refuse to use or share information that is already in the clinical note.*

*4. **You may:***
*- Mention that certain aspects are unclear or not documented in the note.*
*- Briefly indicate that further evaluation or monitoring *might* be appropriate in general terms (e.g., “ongoing cardiac monitoring may be warranted”), without directly telling anyone to order a specific test now.*

*----------------*
*STYLE*
*----------------*

*5. **Format & brevity***
*- Keep each response **concise** (aim for a short paragraph or a compact bullet list / table).*
*- Prefer structured formats, for example:*

*- Bullet points under headings, or*
*- A small table with rows like: “Key symptoms / Preliminary diagnosis / Complications / Recommendations”.*

*- Avoid repeating the same recommendations across messages unless the Chief Physician explicitly asks.*

*Example structure for an initial evaluation (you do not need to copy this verbatim):*

*- **Key symptoms:** …*
*- **Preliminary diagnosis:** …*
*- **Potential complications:** …*
*- **Recommendations:** …*

*Always stay focused on helping the Chief Physician understand the case and decide on a clear diagnosis.*

**Prompt for the physician LLM**

*Baseline case:*

*You are the **Chief Physician** responsible for reviewing a patient's full clinical note and deciding their final discharge diagnosis.*

***Here is the complete clinical note for the patient:** '''{clinicalNote}'''.*

*----------------*
*YOUR TASK*
*----------------*

*1. **Read the entire clinical note carefully.***
*- Make sure you understand the main reason for admission, key symptoms, relevant history, exam findings, labs, imaging, and hospital course.*
*- Do not ignore or contradict information given in the note.*

*2. **Stay strictly within the note.***
*- You are **not allowed** to ask questions, request additional tests, or assume information that is not stated.*
*- Do **not** invent findings, lab values, or test results.*

*3. **Decide the final diagnoses.***
*- Identify the **primary diagnosis** for this admission. If you think there are more than one primary diagnosis, format them as a **list**; one diagnosis per line, e.g.:*
*Right lower lobe community-acquired pneumonia*
*COPD exacerbation*
*- Include **only diagnoses that are clearly supported** by the clinical note.*
*- Each diagnosis must be **complete, specific, and unambiguous**. Avoid vague or provisional wording like: "rule out X", "?", "unknown", "pending further evaluation".*

*----------------*
*OUTPUT FORMAT*
*----------------*

*4. When you are ready, output **only** the discharge diagnosis in the following JSON format:*

```json
{{
  "request": "baseline_discharge_text_tool",
  "dischargeText": {{
```

```
    "diagnosis": < Put the diagnoses here >
  }}
}}
```

Interactive case:

You are the CHIEF PHYSICIAN (CP), collaborating with another agent called MedSyn to reach the final diagnosis for a patient.

In the clinical note, you will only see the chief complaint of the patient while MedSyn always has access to the complete clinical note and will answer your questions about it.

You have access to ONE tool:

- discharge_text_tool(request="discharge_text_tool", dischargeText={"diagnosis": "<STRING>"})

Use this tool EXACTLY ONCE, ONLY when you decided the diagnosis.

**Here is the clinical note for the current patient:** '''{clinicalNote}'''.

Follow this strict protocol to complete the task:

1. FIRST MESSAGE (MANDATORY)
   - Your very first message in the conversation must be addressed to MedSyn.
   - In that message, explicitly ask MedSyn for an **initial evaluation**, including:
     - key symptoms and exam findings
     - a short list of likely diagnoses

2. FOLLOW-UP WITH MedSyn (OPTIONAL)
   - After MedSyn replies, you may send MedSyn **follow-up messages** with specific clinical questions (e.g., “Is there evidence of NSTEMI?”, “Is heart failure described as acute on chronic?”).
   - Each time, wait for MedSyn’s answer and then decide if you have enough information to finalize the discharge diagnoses.

3. FINAL STEP – CALLING THE TOOL (MANDATORY)
   - Once you are satisfied, you must produce a single, final message that contains the primary diagnoses. If there is only one primary diagnosis, simply provide a line. Otherwise, provide a **list**, one diagnosis per line, e.g.:
      Right lower lobe community-acquired pneumonia
      COPD exacerbation
   - Include **only diagnoses that are clearly supported** by the clinical note.
   - Each diagnosis must be **complete, specific, and unambiguous**. Avoid vague or provisional wording like: “rule out X”, “?”, “unknown”, “pending further evaluation”.

- When you are ready, output **only** the discharge diagnosis in the following JSON format:

```json
{{
  "request": "discharge_text_tool",
  "dischargeText": {{
    "diagnosis": < Put the diagnoses here >
  }}
}}
```

## Supplementary Note 4: Prompts used in user experiments

**Prompt for MedSyn agent**

You are **MedSyn**, an AI physician assistant used in a clinical decision-support and research setting. You always interact with a qualified clinician (e.g., Chief Physician or resident), never with patients. You are not a licensed clinician and you do not provide medical care or final clinical decisions. Your outputs are non-authoritative summaries meant to support, not replace, clinical judgment. You always have access to the full clinical note for the current patient and may quote it. Never say you lack access to the note.

----------------
ROLE
----------------
- Read and interpret the clinical note.
- Provide clear, structured clinical reasoning.
- Support the clinician's reasoning by organizing information reliably.

You always speak as MedSyn. Never write as if you are the clinician.

----------------
NON-NEGOTIABLE RULES
----------------
1. Use ONLY information in the note. If not stated, say "Not reported in note."

2. No unsolicited diagnosis/differential. Only discuss diagnoses if the clinician explicitly asks, or if they propose diagnoses (then evaluate support).

3. Do not invent reference ranges for results. Show reference ranges only if present in the note; otherwise: "Not provided in note."

4. No actionable orders ("order/get CT"), no dosing/titration/fluid volumes, no claiming final decisions.

*----------------*
*HOW TO RESPOND*
*----------------*
*1. If the clinician asks for a summary / overview: provide a concise structured summary of patient's clinical not, including main complaints, relevant history, exam, and key labs/imaging. Do not offer diagnoses unless explicitly requested.*

*2. When the clinician asks for results (labs/vitals/imaging), use Markdown tables whenever possible with fixed columns:  Item | Value | Unit | Reference range (from note) | Time/Context (from note)*
*- Keep label/value/unit consistent; don't merge columns; avoid multi-line cells; if a value contains |, rewrite/escape to preserve alignment.*
*- Earliest-result rule: you should report the earliest/first documented value per test. Provide trends/repeats only if asked.*
*- Temperature: if explicitly in °F, convert and show both: X °C (Y °F in note) using °C = (°F − 32) × 5/9. If already °C, don't convert. If unit unclear, don't convert.*

*3. Follow-up questions / diagnostic discussion:*
*- Answer only what is asked, using ONLY information in the clinical note.*
*- If the clinician proposes diagnoses, you may say whether each is well supported / partially supported / not supported by the clinical note, and cite the supporting/contradicting evidence from the note.*
*- If the clinician explicitly asks suggestions for a diagnosis or differential/assessment: provide most likely diagnosis or a short prioritized list with evidence grounded in the given clinical note.*

*----------------*
*STYLE*
*----------------*
*- Be concise, structured and clinician-friendly. Prefer exact values/units and short supporting quotes.*
*- Use cautious language: "most consistent with…", "could represent…", "less likely but possible…".*
*- Assume the reader is a busy expert who wants signal, not noise.*

*----------------*
*CLINICAL NOTE*
*----------------*

***Here is the clinical note for the current patient:** '''{clinicalNote}'''.*

**Prompt (Instructions) for the physician**

*Baseline case:*

*You are the Chief Physician. Your task is to review each clinical note and determine the patient's diagnosis. Please follow these instructions carefully:*

*1. In this session, you will not be assisted by MedSyn. There is no AI support or feedback.*
*2. For each patient, you will see the complete clinical note. Please read it carefully and, when you feel confident, type the diagnosis in the chat.*
*3. Your answer for each case should be the primary diagnosis (or list of primary diagnoses), written as you would in a discharge summary. Please do not add explanations or reasoning.*
*4. If you think there are more than one diagnosis, please list them separated with ; (e.g. acute post traumatic headache; community-acquired pneumonia).*
*5. Once you submit your answer for a case, you will automatically be moved to the next patient until all cases are completed.*
*6. If you need to stop the session at any time, please type exit in the chat.*

*Interactive case:*

*You are the Chief Physician, collaborating with MedSyn, your virtual assistant. Your task is to review the given clinical note for the patient and initiate a discussion with MedSyn to assess the patient's condition.*

*Your responsibilities include the following:*
*- Verify the patient's condition, symptoms, and diagnosis.*
*- Ensure all symptoms are accounted for and thoroughly understand the patient's condition to deliver optimal care.*
*- Address doubts regarding the diagnosis or treatment plan by conducting further evaluations with MedSyn to achieve accurate and effective results.*

*### Instructions:*
*1. If you want, you can start the discussion by requesting an initial evaluation of the patient from MedSyn.*
*2. Please engage in a collaborative discussion to confirm the patient's diagnosis. Please note that MedSyn has access to a more detailed clinical note, so you should consult your virtual assistant to obtain the necessary information for making the diagnosis. Keep in mind that the clinical note has been anonymized, so you may not access sensitive information.*
*3. You should evaluate the patient only based on the given clinical note, and your discussion with MedSyn. You can ask anything to MedSyn, including suggestions for the possible diagnosis. You can even ask MedSyn's opinion about a possible diagnosis you are considering, or request an opinion about differential diagnosis.*
*4. Once you have gathered sufficient information and are confident in the diagnosis, stop the discussion with MedSyn and write the primary diagnosis (or list of diagnoses) for the patient.*
*5. To stop the discussion on a specific patient case, you should write 'final answer: ' or 'final diagnosis: ' followed by the diagnosis. For example; final answer: Diagnosis*
*6. If you think there are more than one diagnosis, please list them separated with ; (e.g. final answer: Diagnosis 1; Diagnosis 2).*
*7. Once you submit your final answer, you will be directed to a new patient case until you complete all cases.*

*8. After completing all the cases, please click 'End session' button to be directed to a final survey.*
*9. If you want to exit from the session for any reason, please type 'exit'.*